\documentclass[letterpaper]{article}
\usepackage{aaai2027}
\nocopyright
\usepackage[hyphens]{url}
\usepackage{graphicx}
\usepackage{natbib}
\usepackage{caption}
\usepackage{algorithm}
\usepackage{algorithmic}
\usepackage{amsmath}
\usepackage{amssymb}
\usepackage{mathtools}
\usepackage{booktabs}
\usepackage{multirow}
\usepackage{xcolor}
\usepackage{alltt}
\title{MASC: A Multi-Agent Self-Calibration Framework with Latent Construct Alignment for Consistent Client Role-Playing in Psychological Counseling}
\author{
Shixin Peng\textsuperscript{\rm 1},
Kun Jiang\textsuperscript{\rm 1},
Jiaxing Zheng\textsuperscript{\rm 1},
Qihao Yang\textsuperscript{\rm 2},
Jingying Chen\textsuperscript{\rm 1,*}
}
\affiliations{
\textsuperscript{\rm 1}Central China Normal University, Wuhan, China\\
\textsuperscript{\rm 2}Shanghai Jiao Tong University, Shanghai, China\\
}

\begin{document}

\maketitle

\begin{abstract}
Large language models are increasingly used to simulate clients for counselor training and psychological counseling research, yet such applications are reliable only when simulated clients remain psychologically coherent across extended interactions. However, existing role-playing methods largely rely on static profile prompting, making them prone to persona drift, unrealistic cooperativeness, and inconsistent psychological states, communicative actions, and emotions. Meanwhile, current evaluations also lack an integrated testbed for both stable client characteristics and evolving psychological dynamics. To address these issues, we propose \textbf{MASC}, a Multi-Agent Self-Calibration framework with latent construct alignment, for consistent client role-playing in psychological counseling conversation. MASC forms a calibration loop that integrates construct-guided generation, collaborative refinement, consistency verification, and memory-based revision, enabling role inconsistencies to be detected and corrected as the dialogue unfolds. To define the calibration targets and evaluate them systematically, we further introduce \textbf{CRPC-Bench}, a Client Role-Playing Consistency Benchmark covering session-level profile information and Big-Five personality traits, together with turn-level psychological state, communicative action, and emotion expression. CRPC-Bench is instantiated on 38 motivational interviewing client profiles with additional personality and emotion annotations. Experiments show that MASC consistently outperforms existing methods across profile, personality, receptivity, and turn-level consistency, with the heterogeneous configuration achieving the strongest overall performance. Together, MASC and CRPC-Bench provide a unified foundation for developing and evaluating psychologically coherent client simulations for AI-assisted counseling research and training. Code and data are available at \url{https://github.com/DeSun99/MASC-CRPC}.

\end{abstract}

\section{Introduction}
Large language models are increasingly used to simulate clients for counselor training and psychological counseling research \cite{park2023generativeagents}. Such systems can provide scalable and repeatable interactions across diverse client profiles, reducing the dependence on real clients in training and experimental settings. Their value, however, depends on whether the simulated client remains psychologically coherent throughout an extended counseling session.

Psychological counseling requires more than fluent and locally plausible responses. A simulated client must preserve stable background information, motivations, beliefs, and personality traits, while showing appropriate changes in psychological state, communicative action, and emotion. This is particularly important in motivational interviewing, where resistance, hesitation, and gradual readiness for change are meaningful parts of the counseling process \cite{miller2013motivational}. General purpose LLMs, which are typically optimized to be helpful, may instead become unrealistically cooperative, reveal change plans prematurely, or drift from the assigned client role.

Existing work has evaluated persona consistency, personality expression, memory preservation, and profile dialogue alignment in general role-playing settings \cite{yuan2025dmtrolebench}. In mental health research, simulated clients and digital patients have also been developed for counselor training, while recent methods model client profiles and psychological state transitions. However, existing generation methods largely rely on static profiles and lack turn by turn mechanisms for detecting and correcting role inconsistencies\cite{LI2025108687,chen2025mind,lee2024cactuspsychologicalcounselingconversations}. Current evaluations also lack a unified testbed that jointly assesses stable client characteristics and evolving psychological dynamics.

In this paper, we investigate an important yet underexplored question: \textit{how can LLMs be systematically calibrated and evaluated for psychologically coherent client role-playing in counseling conversations?} This problem presents two main challenges. The first is to establish an adaptive calibration mechanism that identifies and corrects construct mismatches as the dialogue unfolds. The second is to develop an integrated evaluation foundation that captures consistency both across the full session and within individual dialogue turns.

To address the first challenge, we propose \textbf{MASC}, a Multi-Agent Self-Calibration framework with latent construct alignment. MASC integrates construct based generation, collaborative refinement, consistency verification, and memory based revision into a closed calibration loop. It uses expected psychological constructs as generation and verification targets, allowing inconsistent responses to be revised before they enter the dialogue history.

To address the second challenge, we introduce \textbf{CRPC-Bench}, a Client Role-Playing Consistency Benchmark that defines the calibration targets and provides a unified testbed. It evaluates profile information and Big-Five personality traits at the session-level, together with psychological state, communicative action, and emotion expression at the turn-level. We instantiate CRPC-Bench on 38 motivational interviewing client profiles and augment the original dataset with personality and emotion annotations. CRPC-Bench therefore specifies what psychological coherence requires, while MASC uses the same constructs to calibrate response generation.

Experiments show that MASC consistently improves profile, personality, receptivity, and turn-level consistency over existing methods. The heterogeneous configuration achieves the strongest overall performance. Ablation results further show that the complete generation, verification, and retry loop yields the most balanced performance, while controlled removals clarify the complementary roles of its components. 

Our main contributions are summarized as follows:

\begin{itemize}

\item We propose \textbf{MASC}, a Multi-Agent Self-Calibration framework with latent construct alignment that detects and corrects client role inconsistencies during multi-turn counseling conversations.

\item We introduce \textbf{CRPC-Bench}, a benchmark that jointly evaluates stable client characteristics at the session-level and evolving psychological dynamics at the turn-level.

\item We conduct extensive experiments comparing homogeneous and heterogeneous MASC configurations and examining the complementary roles of candidate refinement, construct verification, and retry memory.

\end{itemize}

\section{Related Work}

\textbf{LLM Role-Playing and Mental Health.} Recent work has examined whether LLMs can sustain personalized and role-consistent behavior in dialogue. PersonaLLM shows that LLMs can express personality traits, but such expression is unstable across contexts \cite{jiang-etal-2024-personallm}. CharacterEval and persona-aware contrastive learning further emphasize explicit persona consistency beyond surface fluency \cite{tu-etal-2024-charactereval,ji-etal-2025-enhancing}. In mental-health dialogue, this challenge becomes more demanding because client simulators must preserve profiles, motivations, psychological states, and emotional responses across counseling turns \cite{yang2025consistentclientsimulationmotivational}. Recent studies have explored LLM role-playing for mental-health applications \cite{LI2025108687,chen2025mind,lee2024cactuspsychologicalcounselingconversations}: MELT supports emotion-oriented annotation for affective modeling \cite{jing2025melt}; PsychAdapter adapts LLMs to personality, demographic, and mental-health characteristics \cite{vu2025psychadapteradaptingllmtransformers}; Roleplay-doh enables senior mental-health supporters to create customized AI patients for novice counselor training \cite{louie-etal-2024-roleplay}; and PATIENT-$\Psi$ provides a patient simulation framework for cognitive behavioral therapy training  \cite{wang-etal-2024-patient}. However, these methods largely condition generation on static profiles and focus on general persona fidelity; none of them jointly maintains session-level personality traits and turn-level psychological dynamics, nor do they provide a turn-by-turn mechanism to detect and correct role inconsistencies as they emerge. Our work addresses precisely this gap through construct-aligned self-calibration.

\noindent\textbf{Consistency Evaluation of LLM Role-Playing.} A complementary line of work develops benchmarks and protocols for evaluating role-playing consistency \cite{liu2025surveypersonalizedlargelanguage,xiao-etal-2025-towards,xie2024largelanguagemodelagents}. RoleMRC \cite{lu-etal-2025-rolemrc} introduced a fine-grained benchmark for role-playing and instruction following. Beyond Dialogue \cite{yu-etal-2025-beyond} improves sentence-level profile-dialogue alignment across diverse role-playing scenarios. PsyMem \cite{cheng2025psymemfinegrainedpsychologicalalignment} integrates fine-grained psychological attributes and explicit memory control to improve memory consistency. RoleLLM \cite{wang2024rolellmbenchmarkingelicitingenhancing} further benchmarks, elicits, and enhances role-playing abilities in LLMs. Although these studies define diverse benchmarks and evaluation protocols, direct extension to mental health counseling remains insufficient because they typically lack psychological-theory-grounded guidance and show limited generalization across counseling scenarios.

\begin{figure*}[t]
  \centering
  \includegraphics[width=0.95\textwidth]{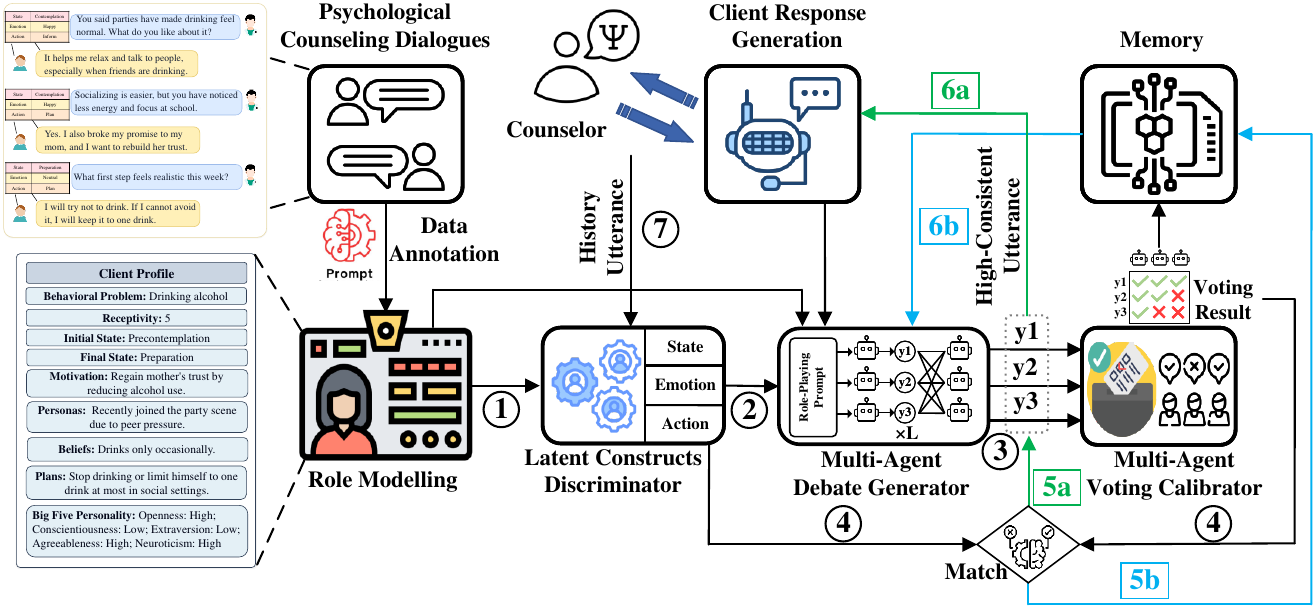}
  \caption{The proposed Multi-Agent Self-Calibration framework with latent construct alignment. The framework combines latent construct prediction, multi-agent debate, voting calibration, and memory feedback to generate consistent client responses.}
  \label{fig:overview}
\end{figure*}

\section{A Multi-Agent Self-Calibration Framework}

At turn $n$, the framework observes the generated history and current counselor utterance, $H_n=\{(Q'_t,R'_t)\}_{t=1}^{n-1}\cup\{Q'_n\}$. It does not use future client responses or a complete reference dialogue during generation. Session-level constraints $C^{sess}=[PI,BFT]$ describe profile information and Big-Five personality traits that should persist across the dialogue. Turn-level guidance $z_n=[EE_n,PS_n,CA_n]$ specifies the emotion expression, psychological state, and communicative action expected in the current response $R'_n$. Thus, session-level constructs constrain who the client remains, while turn-level constructs guide how the client responds at the current point in the interaction.

MASC generates and verifies each response through a self-calibration loop. The first attempt is generated from the dialogue context and target constructs. If the voted labels of the response do not match the target constructs, retry memory records the mismatch and conditions the next generation attempt:
\begin{equation}
R'_{n,r} =
\begin{cases}
\mathcal{G}(H_n,C^{sess},z_n;\mathcal{M}), & r=0,\\
\mathcal{G}(H_n,C^{sess},z_n;\mathcal{M},\mathcal{B}_{n,r}), & r>0,
\end{cases}
\end{equation}
% Plan: Define the roles of the agent sets and connect the formal loop to the operational modules.
where $r$ is the retry index, $\mathcal{M}$ contains generation agents $\mathcal{M}_g$ and judge agents $\mathcal{M}_j$, and $\mathcal{B}_{n,r}$ stores failed attempts together with construct-level mismatch feedback. Figure~\ref{fig:overview} and Algorithm~\ref{alg:masc-generation} show how the latent constructs discriminator, debate generator, voting calibrator, and retry memory form a closed loop. The exact role-playing, refinement, annotation, and memory templates are documented in Appendix~\ref{app:prompt-templates}; the method description below specifies the information passed between these templates.

\subsection{Latent Constructs Discriminator}

The latent constructs discriminator determines the target control signal before the client response is generated. This step makes the generation process explicit: instead of relying only on the prompt to implicitly preserve the role, MASC first predicts the emotion, psychological state, and communicative action that the next client response should express. The label spaces are summarized in Figure~\ref{fig:metrics-system}; we denote them as $\mathcal{C}_{EE}$, $\mathcal{C}_{PS}$, and $\mathcal{C}_{CA}$:
\begin{align*}
\mathcal{C}_{EE}={}&\{\text{neutral},\text{happy},\text{sad},\text{angry},
\text{fearful},\text{disgusted},\text{surprised}\},\\
\mathcal{C}_{PS}={}&\{\text{Precontemplation},\text{Contemplation},
\text{Preparation}\},\\
\mathcal{C}_{CA}={}&\{\text{Deny},\text{Downplay},\text{Blame},\text{Hesitate},
\text{Doubt},\text{Engage},\\
&\quad\text{Inform},\text{Acknowledge},\text{Accept},\text{Reject},
\text{Plan},\text{Terminate}\}.
\end{align*}

For emotion expression, the discriminator uses the recent dialogue context to select the most likely emotion label. This prediction provides the affective target for the next client response:
\begin{equation}
EE_n = \arg\max_{e\in \mathcal{C}_{EE}} p_{\theta}(e\mid H_n).
\end{equation}

For psychological state, we model the client as moving through counseling-relevant stages. The state changes only when the counselor addresses the client's motivation and when belief barriers are sufficiently resolved. Let $T_n$ be the top-5 topics predicted from the recent dialogue context, $t^\star$ the client's target motivation topic, $m_n\in\{0,1\}$ an indicator of whether the counselor utterance matches the target motivation, and $\mathcal{B}^{belief}_n$ the set of unresolved belief barriers. The transition rule is
{\footnotesize
\begin{equation}
\vcenter{\hbox{$%
\begin{aligned}
PS_n =
\left\{
\begin{alignedat}{2}
&\text{Contemplation}, && PS_{n-1}=\text{Precontemplation},\\[-1pt]
& && t^\star\in T_n,\ m_n=1,\\[1pt]
&\text{Preparation}, && PS_{n-1}=\text{Contemplation},\\[-1pt]
& && |\mathcal{B}^{\mathrm{belief}}_n|=0,\\[1pt]
&PS_{n-1}, && \text{otherwise}.
\end{alignedat}
\right.
\end{aligned}
$}}
\end{equation}
}

For communicative action, the discriminator combines two sources of evidence. The context-dependent term identifies actions that are locally plausible after the counselor's latest utterance, while the receptivity-aware prior discourages a low-receptivity client from moving prematurely toward acknowledgement, acceptance, or planning:
\begin{equation}
p^{ctx}_n(a)=p_{\theta}(a\mid H_n), \quad p^{rec}_n(a)=p(a\mid \rho),
\end{equation}
where $\rho\in\{1,\ldots,5\}$ denotes the client receptivity level. The combined scores are normalized over the action label space:
\begin{equation}
p_n(a) \propto p^{ctx}_n(a) + p^{rec}_n(a),
\end{equation}
and the communicative action is sampled from this distribution:
\begin{equation}
CA_n\sim p_n(a).
\end{equation}
Together, these three predictions form the turn-level target $z_n=[EE_n,PS_n,CA_n]$ used by the generator and calibrator.

% Plan: Explain how the three mathematical targets are operationalized as distinct constrained prediction tasks.
Operationally, MASC uses a separate constrained annotation template for each turn-level dimension. Each template receives the recent dialogue context and its dimension-specific label set, and returns one discrete label rather than a free-form assessment. Separating emotion, psychological state, and communicative action prevents a salient cue in one dimension from implicitly determining the other two; the resulting labels are assembled into $z_n$ only after the three predictions are completed.

\subsection{Multi-Agent Debate Generator}

Given the dialogue context, session-level constructs, and turn-level target labels, the generator produces multiple candidate client responses through multi-agent debate\cite{du2023improving}. Within each attempt, the debate process has two related stages. In the first debate step, the generation agents independently produce different candidate responses under the same conditions, expanding the candidate space. In each later debate step, every agent observes the previous-step responses of the other agents and revises its own candidate. This peer refinement incorporates complementary information before the final candidate set is returned.

\begin{algorithm}[t]
\caption{MASC Self-Calibration at Turn $n$}
\label{alg:masc-generation}
\begin{algorithmic}[1]
\STATE \textbf{Input:} Dialogue context $H_n$, session constructs $C^{sess}$, target constructs $z_n$, generation agents $\mathcal{M}_g$, judge agents $\mathcal{M}_j$, initial retry memory $\mathcal{B}_{n,0}=\emptyset$
\STATE \textbf{Output:} Final client response $R_n^\star$
\FOR{each attempt $r\in\{0,1,2\}$}
    \STATE Generate $\mathcal{Y}_{n,r}$ through multi-agent debate using $H_n$, $C^{sess}$, $z_n$, and $\mathcal{B}_{n,r}$
    \STATE Use $\mathcal{M}_j$ to obtain voted labels and agreement scores for every candidate
    \STATE Select $R^\star_{n,r}$ and identify its mismatched dimensions $\Delta_{n,r}$
    \IF{$R^\star_{n,r}$ has sufficient voting agreement and $\Delta_{n,r}=\emptyset$}
        \STATE \textbf{return} $R^\star_{n,r}$
    \ELSE
        \STATE Construct a failure record from $R^\star_{n,r}$, $z_n$, and its voted labels
        \STATE Update $\mathcal{B}_{n,r+1}$ while retaining at most the two latest records
    \ENDIF
\ENDFOR
\STATE \textbf{return} the best-matching candidate response
\end{algorithmic}
\end{algorithm}

Let $\mathcal{M}_g=\{M_1,M_2,\ldots,M_K\}$ be the set of generation agents. In our implementation, $K=3$ and each attempt contains $L=5$ debate steps. This fixed budget allows agents to revise inconsistent response cues while keeping generation cost bounded.

In the first debate step, each agent independently generates a response conditioned on the same dialogue context and target constructs:
\begin{equation}
R^{(1)}_{n,i}\sim \pi_{\theta_i}
\left(
\cdot \mid H_n,C^{sess},z_n
\right),
\quad i=1,\ldots,K.
\end{equation}
For later debate steps, each agent observes the previous-step responses from the other agents. When retry memory is available, the same generation step also conditions on failed attempts from earlier attempts, helping the agent avoid repeated inconsistencies:
\begin{equation}
R^{(l)}_{n,i}\sim
\pi_{\theta_i}
\left(
\cdot \mid H_n,C^{sess},z_n,
\{R^{(l-1)}_{n,j}\}_{j\neq i},
\mathcal{B}_{n,r}
\right).
\end{equation}
After $L$ debate steps, the generator returns the final candidate set
\begin{equation}
\mathcal{Y}_n=\{R^{(L)}_{n,1},R^{(L)}_{n,2},\ldots,R^{(L)}_{n,K}\}.
\end{equation}

% Plan: Clarify which prompt fields implement the conditioning variables in the debate equations.
The debate prompt instantiates these conditioning variables explicitly. Stable role constraints---behavioral problem, counseling goal, profile, reference style, and optional Big-Five traits---remain fixed across the session. At each debate step, the prompt adds $H_n$ and the target labels $z_n$; later steps additionally include peer responses from the preceding step and, for a retry attempt, the rendered records in $\mathcal{B}_{n,r}$. Agents are required to output only one client utterance, ensuring that the candidates passed to the calibrator are comparable rather than accompanied by rationales or metadata. Detailed debate pseudocode is provided in Appendix~\ref{app:detailed-algorithms}.

\subsection{Multi-Agent Voting Calibrator}
Inspired by the LLM-as-a-Judge paradigm, the voting calibrator evaluates whether each candidate response expresses the expected latent constructs. Rather than relying on a single judge, MASC asks multiple judge agents $\mathcal{M}_j=\{M_1,M_2,\ldots,M_K\}$ to annotate each candidate and aggregates their labels by majority voting. For candidate $R^{(L)}_{n,i}$, judge agent $M_j$ predicts one label for each latent dimension:
\begin{equation}
\hat{z}^{d}_{n,i,j}
=
D^{d}_{\theta_j}(H_n,R^{(L)}_{n,i}),
\quad d\in\{EE,PS,CA\}.
\end{equation}
For each dimension, the majority-voted label is computed as
\begin{equation}
v^d_{n,i}
=
\arg\max_{c\in\mathcal{C}_d}
\sum_{j=1}^{K}\mathbb{I}
\left(\hat{z}^{d}_{n,i,j}=c\right),
\end{equation}
where $\mathcal{C}_d$ is the label space of dimension $d$.

The calibrator then checks two distinct properties. First, judge agreement filters out candidates whose inferred construct labels are ambiguous across judges. This asks whether the judges interpret a candidate consistently, not whether their interpretation is the desired one:
\begin{equation}
\gamma^d_{n,i}
=
\mathbb{I}
\left[
\max_{c\in\mathcal{C}_d}
\sum_{j=1}^{K}\mathbb{I}
(\hat{z}^{d}_{n,i,j}=c)
\geq
\mathrm{round}(\tau K)
\right],
\end{equation}
where $\tau=2/3$, requiring at least two of three judges to agree. The overall agreement score is
\begin{equation}
S^{agree}_{n,i}
=
\sum_{d\in\{EE,PS,CA\}}\gamma^d_{n,i}.
\end{equation}
Second, target matching checks whether the majority-voted label equals the target predicted by the discriminator. Thus, judges may agree strongly on a candidate that nevertheless expresses the wrong construct:
\begin{equation}
S^{match}_{n,i}
=
\sum_{d\in\{EE,PS,CA\}}
\mathbb{I}
\left(v^d_{n,i}=z^d_n\right).
\end{equation}
The selected response prioritizes reliable voting agreement and then target-label alignment:
\begin{equation}
R^\star_n
=
\arg\max_{R^{(L)}_{n,i}\in\mathcal{Y}_n}
\left(S^{agree}_{n,i},S^{match}_{n,i}\right).
\end{equation}
% Plan: Connect majority voting to the actual annotation protocol and the accept-or-retry decision.
The calibrator implements this scoring rule with three dimension-specific annotation prompts. Judge agents independently return EE, PS, and CA labels; majority voting is performed within each dimension before the three voted labels are compared with $z_n$. The calibrator outputs the selected response, its majority-voted labels, and the mismatch set $\Delta_{n,r}$. A fully matched candidate follows Step~5a in Figure~\ref{fig:overview} and enters the dialogue history. Otherwise, Step~5b passes these outputs to Retry Memory, which organizes them into a failure record for the next attempt. Thus, Voting supplies the construct-level diagnosis, whereas Retry Memory is responsible for storing and transferring it. Detailed voting pseudocode is provided in Appendix~\ref{app:detailed-algorithms}.
\subsection{Retry Memory}

% Plan: Specify the contents and use of retry memory as construct-level corrective feedback.
Retry Memory provides targeted feedback when no candidate satisfies all three constructs. For a failed attempt, the response text $R^\star_{n,r}$ originates from the Debate Generator, while its majority-voted labels $v_{n,r}$ and mismatch set $\Delta_{n,r}$ are supplied by the Voting Calibrator. Retry Memory then renders each mismatch in $\Delta_{n,r}$ as an expected-versus-actual explanation $\epsilon_{n,r}$ and organizes the information as
\begin{equation}
b_{n,r}=(r,R^\star_{n,r},z_n,v_{n,r},\epsilon_{n,r}).
\end{equation}
The record therefore tells the next debate both what was generated and which constructs failed to match. Before the next attempt, the record is appended to the debate prompt so that the agents receive a concrete correction target rather than a generic instruction to improve consistency. Detailed memory-update pseudocode is provided in Appendix~\ref{app:detailed-algorithms}.

% Plan: Clarify the retry budget without replacing the existing retry mechanism.
For each client response, MASC makes at most three attempts. The first is the initial generation without retry memory, and the next two are retries guided by previous failure records. Each attempt contains five debate steps. Retry Memory retains at most the two most recent failures ($B=2$), which bounds the prompt length while preserving the feedback needed for both retries. A response is accepted when the judges reach sufficient agreement and all majority-voted labels match $z_n$; after the three attempts are exhausted, the best-matching candidate observed across the attempts is returned to preserve dialogue continuity.

\section{Client Role-Playing Consistency Benchmark}\label{SLSL Metrics}
\begin{figure}[tbp]
  \centering
  \includegraphics[width=0.37\textwidth]{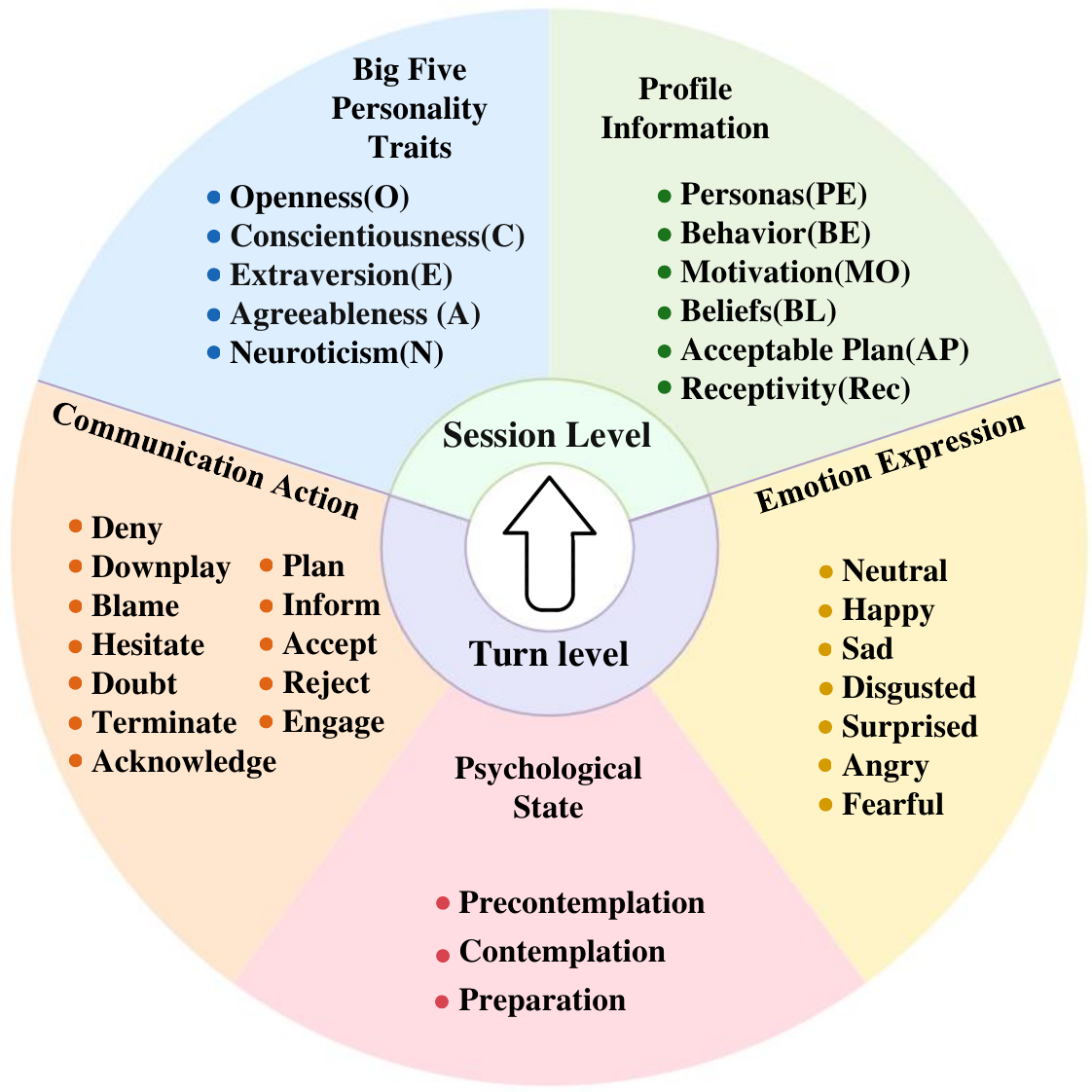}
  \caption{CRPC-Bench dimensions for client role-playing consistency, covering session-level profile information and Big-Five personality traits, and turn-level psychological state, communicative action, and emotion expression.}
  \label{fig:metrics-system}
\end{figure}

To provide MASC with calibration targets and a systematic testbed, we construct CRPC-Bench in consultation with counseling psychology experts. Its dimensions draw on machine psychology \cite{hagendorff2023machine}, the Big-Five taxonomy \citep{John1999}, the transtheoretical model \citep{prochaska1997transtheoretical}, and Ekman's basic emotions \citep{ekman1992there}. As shown in Figure~\ref{fig:metrics-system}, session-level constructs assess whether a complete dialogue preserves the client's profile and enduring personality, whereas turn-level constructs assess psychological state, communicative action, and emotion as they evolve. The two levels capture different failure modes: a dialogue may preserve profile facts while moving unrealistically quickly toward change, or produce plausible local reactions while gradually contradicting the assigned role. Evaluating both levels is therefore necessary.

\subsection{Session-Level Latent Constructs}
Session-level constructs are evaluated over the complete generated session. We consider profile information, receptivity dynamics, and Big-Five personality consistency. The original MI dataset provides structured profiles and receptivity labels \cite{yang2025consistentclientsimulationmotivational}; we additionally annotate each client with High/Low labels on the five BFI dimensions \citep{John1999}.

\begin{itemize}
\item \textbf{Profile Information}: 
Following \citet{yang2025consistentclientsimulationmotivational}, each profile specifies the behavioral problem, persona, motivation, beliefs, acceptable change plans, five-level receptivity, and the initial/final stage of change. Persona records relevant background facts, motivation identifies reasons that may support change, beliefs capture barriers or concerns, and acceptable plans delimit changes the client could plausibly endorse. Evaluating these fields over the complete session can reveal contradictions that are not apparent in an isolated response.

\item \textbf{Big-Five Personality Traits}: We assess whether a dialogue expresses the instructed openness, conscientiousness, extraversion, agreeableness, and neuroticism rather than merely repeating trait labels. Evidence is drawn from behavior across the dialogue, including organization, social expressiveness, reactions to uncertainty, and concern for others. Our annotation protocol reaches 78.92\% accuracy on PersonalityEvd~\citep{sun2024revealing} and approximately 87.5\% inter-annotator agreement on a random 30\% benchmark sample independently annotated by two experts and two trained graduate students.

\end{itemize}

\subsection{Turn-Level Latent Constructs}
Turn-level constructs cover Psychological State, Communicative Action, and Emotion Expression for each client response. Unlike session-level traits, they may change as the counselor introduces information and the client reacts. The original MI dataset provides state and action annotations but lacks utterance-level emotion labels. We annotate emotions with \texttt{ChatGPT-5.5}~\cite{openai2023gpt4technicalreport}; the same prompt achieves 76.12\% accuracy on a balanced seven-class DailyDialog sample~\cite{li-etal-2017-dailydialog}.

\begin{itemize}
\item \textbf{Psychological State}: 
We use the transtheoretical model of health behavior change to define three counseling-relevant client states: \emph{Precontemplation}, \emph{Contemplation}, and \emph{Preparation}~\cite{samhsa2019enhancing,prochaska1997transtheoretical,hashemzadeh2019transtheoretical}. These states capture the client's progression from resistance to readiness for change, which matches the focus of motivational interviewing and the distribution of states in our dataset.

\item \textbf{Communicative Action}:
Inspired by~\cite{dutt-etal-2021-resper}, we tailor actions to client utterance functions in counseling. Labels range from resistance (e.g., Deny and Blame), through information and ambivalence (e.g., Inform and Hesitate), to change-oriented responses (e.g., Acknowledge and Plan). They describe conversational function rather than state or emotion.

\item \textbf{Emotion Expression}: Each client utterance is assigned one of seven emotion labels: Happy, Sad, Fearful, Angry, Surprised, Disgusted, or Neutral. These labels support turn-level evaluation of whether generated clients preserve realistic affective patterns.

\end{itemize}

These turn-level labels remain separate because they describe complementary aspects of one response. A client may, for example, remain in Precontemplation, use an Inform action to describe current behavior, and express fear about its consequences. Emotional concern should not automatically be interpreted as readiness for behavioral change.

\subsection{Consistency Assessment}
Together, these dimensions assess stable characteristics over a session and dynamic constructs at each turn. Within MASC, turn-level labels guide response selection and revision. After generation, a separate evaluator relabels the complete outputs of every method under the same protocol; the resulting distributions are compared with real dialogues using KL divergence. This separation keeps generation guidance distinct from experimental measurement: MASC's internal voting labels are not reused as its reported scores.

\section{Experiment}

\subsection{Experimental Setup}

The experiments are conducted on the MI counseling dialogue dataset introduced by~\cite{yang2025consistentclientsimulationmotivational}, which contains 38 annotated counselor--client dialogue profiles. Each profile includes the client's behavioral problem, initial and final state-of-mind, persona/background information, motivation, belief barriers, acceptable plans, and receptivity level. We further augment the original dataset with Big-Five personality traits and utterance-level emotion expression labels to support session-level personality evaluation and turn-level affective consistency assessment. For the main experiments, \textbf{MASC(Homo)} uses the Qwen-series ensemble (\texttt{Qwen-plus}, \texttt{Qwen-turbo}, and \texttt{Qwen-max})~\cite{qwen2023}, whereas \textbf{MASC(Hetero)} uses \texttt{ChatGPT-5.5}, \texttt{Minimax-M3}~\cite{lai2026minimaxsparseattention}, and \texttt{Deepseek-v4-pro}~\cite{deepseekai2026deepseekv4}; ablations use the homo setting as the Full Model.

For counselor-side interaction, we adopt the counselor-agent prompt from CSF~\cite{yang2025consistentclientsimulationmotivational}, which follows MI-oriented prompting validated in prior studies~\cite{chiu-etal-2024-computational,wang-etal-2024-client-centered,yosef-etal-2024-assessing}. This established prompt has been shown to support therapeutic alliance and counseling skills, and is reused as our counselor-side protocol. We compare MASC with seven baselines: \textbf{Base} uses only the client behavioral problem to prompt response generation~\cite{deng-etal-2023-prompting}; \textbf{Example-Based} conditions generation on a real counseling-session exemplar~\cite{chiu-etal-2024-computational}; \textbf{Profile-Based} uses the structured client profile~\cite{yosef-etal-2024-assessing,wang-etal-2024-client-centered}; \textbf{ProAct-Based} adds descriptions of possible client actions~\cite{zhang-etal-2024-strength}; \textbf{Pro+Dial-Based} further includes dialogue context in the prompt~\cite{shanahan2023role}; \textbf{CSF} denotes the consistent client simulation framework~\cite{yang2025consistentclientsimulationmotivational}; and \textbf{Schedule-ST} follows a fixed state-transition schedule, assigning turns 1--10, 11--20, and 21--30 to Precontemplation, Contemplation, and Preparation, respectively.

% Plan: State the generation and calibration budgets needed to interpret the ablation and convergence results.
For each client turn, MASC uses three generation agents and five debate steps per attempt, with one initial attempt and at most two retries. Candidate responses are annotated separately for EE, PS, and CA by the judge agents, then ranked by voting agreement and target-label match. Failed attempts carry their construct-level mismatch records into the next retry. The full prompt schemas and output constraints are provided in Appendix~\ref{app:prompt-templates}; all ablations retain the same three-Qwen backbone setting and change only the component specified by the variant name.

The separate evaluator uses the same prompts and label definitions for real dialogues, all baselines, and MASC variants. Method names and generation sources are hidden, and samples are presented without method-specific cues. Thus, an internally accepted MASC response must still be relabeled under the common evaluation protocol.

\subsection{Session-level Consistency}

\textbf{Profile and receptivity consistency.}
We jointly evaluate whether generated sessions preserve stable client profiles and realistic receptivity dynamics.

\begin{table*}[!t]
\centering
\small
\setlength{\tabcolsep}{3pt}
\begin{tabular*}{\textwidth}{@{\extracolsep{\fill}}lcccccccc@{}}
\toprule
\multirow{2}{*}{\textbf{Method}}
    & \multicolumn{5}{c}{\textbf{Profile Consistency (\%)}} 
    & \multicolumn{3}{c}{\textbf{Receptivity Consistency}} \\
\cmidrule(lr){2-6} \cmidrule(lr){7-9}
& \textbf{PE} & \textbf{BE} & \textbf{MO} & \textbf{BL} & \textbf{AP}
& \textbf{Avg Rec} & \textbf{MR@20 (\%)} & \textbf{Avg MS} $\uparrow$ \\
\midrule
Base                    & 20.18 & 27.48 & 25.44 & 22.81 & 17.54 & 4.59 & 100.0 & 6.43 \\
Example-Based           & 77.19 & 78.26 & 76.32 & 46.49 & 34.21 & 4.78 & 100.0 & 7.54 \\
Profile-Based           & 69.38 & 79.45 & 71.52 & 56.37 & 47.37 & 4.83 & 95.02 & 10.23 \\
ProAct-Based            & 73.35 & 82.01 & 73.21 & 61.03 & 38.60 & 4.70 & 96.12 & 8.85 \\
Pro+Dial-Based          & 71.35 & 86.49 & 74.21 & 43.51 & 59.46 & 4.73 & 94.32 & 9.47 \\
CSF                     & 70.57 & 81.80 & 73.37 & 71.70 & \textbf{68.51} & 3.45 & 68.03 & 15.73 \\
Schedule-ST             & 71.05 & 89.47 & 50.00 & 57.89 & 55.26 & 3.67 & 94.12 & 11.68 \\
\midrule
\textbf{MASC(Homo)} & 76.32 & 92.11 & 76.84 & 75.79 & 64.74 & 3.41 & 47.47 & 16.39 \\
\textbf{MASC(Hetero)} & \textbf{77.68} & \textbf{98.71} & \textbf{79.74} & \textbf{78.95} & 65.91 & \textbf{3.34} & \textbf{47.13} & \textbf{19.55} \\
\midrule
Real & 100.0 &100.0 &100.0 &100.0 &100.0 & 3.27 & 48.39 & 27.56 \\
\bottomrule
\end{tabular*}
\caption{Session-level consistency results. Left: profile consistency across five attributes, where higher values indicate stronger alignment with the original client profile. Right: receptivity consistency compared with real human dialogues (Real). Receptivity consistency is measured by average receptivity (Avg Rec),  receptivity variation rate within the first 20 turns (MR@20\%), and average maximum stage (Avg MS).}
\label{tab:overall-consistency}
\end{table*} 
Table~\ref{tab:overall-consistency} shows that the Base method is weak across the five profile attributes, suggesting that the behavioral problem alone does not sustain a detailed role. Profile- and action-based baselines improve particular fields, but the gains are uneven. \textbf{MASC(Hetero)} provides the strongest overall profile preservation, leading on Personas, Behavior, Motivation, and Beliefs. Acceptable Plans is the exception: CSF obtains 68.51 versus 65.91 for MASC(Hetero). This exception identifies plan preservation as a remaining error rather than supporting uniform dominance across attributes. On receptivity, MASC(Hetero) is close to Real in Avg Rec (3.34 vs.\ 3.27) and MR@20 (47.13\% vs.\ 48.39\%). Its Avg MS of 19.55 is the best simulated result but remains below the real-dialogue value of 27.56. The receptivity metrics are interpreted relative to Real rather than as independent ``higher is better'' or ``lower is better'' targets. Overall, MASC most closely reproduces the observed pattern without matching the full depth of progression in real sessions.

% Plan: Delimit text-overlap metrics as auxiliary because valid generated sessions may follow different scenarios.
Because a generated counseling session can remain role-consistent while following a different concrete scenario or conversational path from its reference, surface-overlap metrics are not treated as primary evidence in the main evaluation. We report BLEU, ROUGE-L, METEOR, and lexical-diversity results as auxiliary diagnostics in Appendix~\ref{app:semantic-analysis}.

\noindent\textbf{Big-Five personality traits consistency.}
We further evaluate whether simulated clients preserve their original Big-Five personality tendencies across the generated session. The overall score averages accuracy over openness, conscientiousness, extraversion, agreeableness, and neuroticism.

\begin{figure}[t]
  \centering
  \includegraphics[width=0.37\textwidth]{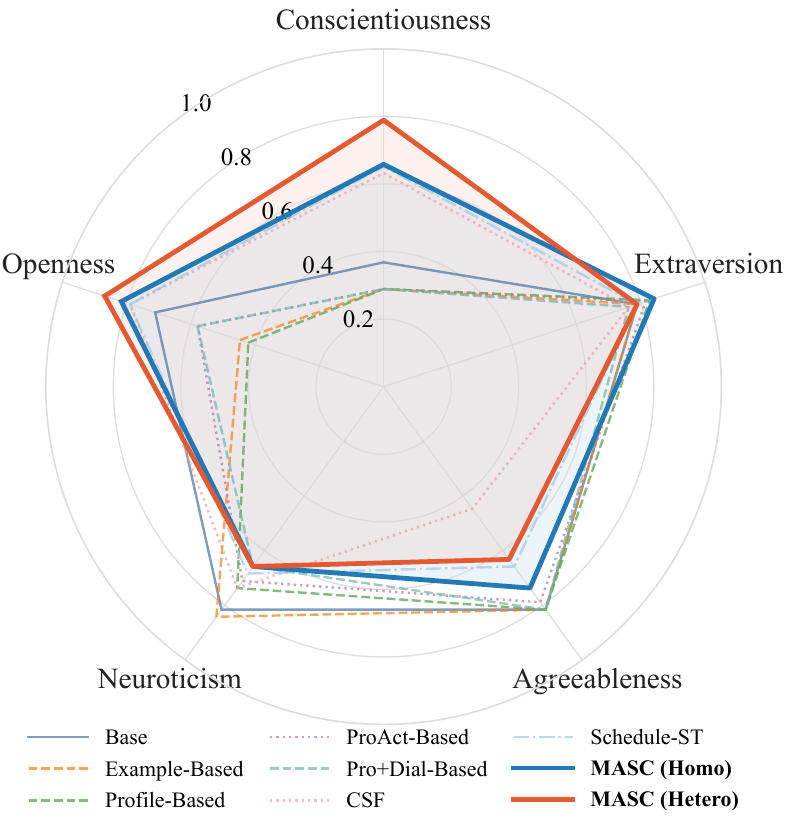}
  \caption{Big-Five personality traits consistency. Radar plot comparing how well each method preserves the five personality dimensions; higher scores indicate stronger alignment with the original client profiles.}
  \label{fig:big-five-consistency}
\end{figure}

As shown in Figure~\ref{fig:big-five-consistency}, \textbf{MASC(Hetero)} achieves the highest overall Big-Five consistency and leads on openness and conscientiousness, whereas \textbf{MASC(Homo)} leads on extraversion. Thus, Hetero is strongest in aggregate but does not dominate every trait. Because the settings also use different model families, this comparison does not isolate model diversity as the cause.

\subsection{Turn-level Consistency}
For each turn-level dimension, we compute KL divergence between the label distributions of a generated method and real client dialogues. Lower values indicate closer corpus-level frequencies and complement response-level target matching: a method may satisfy many local targets while still overproducing a small set of actions or emotions across the dataset.

\begin{table}[t]
\centering
\small
\setlength{\tabcolsep}{2pt}
\begin{tabular*}{\columnwidth}{@{\extracolsep{\fill}}lccc@{}}
\toprule
\textbf{Method} & \textbf{Action KL} $\downarrow$ & \textbf{Emotion KL} $\downarrow$ & \textbf{State KL} $\downarrow$ \\
\midrule
Base & 0.41 & 0.51 & 0.59 \\
Example-Based & 0.23 & 0.77 & 0.43 \\
Profile-Based & 0.17 & 0.63 & 0.45 \\
ProAct-Based & 0.15 & 0.62 & 0.38 \\
Pro+Dial-Based & 0.15 & 0.57 & 0.35 \\
CSF & 0.07 & 0.56 & 0.13 \\
Schedule-ST & 0.10 & 0.33 & 0.18 \\
\midrule
\addlinespace[1pt]
\textbf{MASC(Homo)} & \textbf{0.06} & 0.29 & 0.07 \\
\textbf{MASC(Hetero)} & \textbf{0.06} & \textbf{0.27} & \textbf{0.06} \\
\midrule
\addlinespace[1pt]
Real & 0.00 & 0.00 & 0.00 \\
\bottomrule
\end{tabular*}
\caption{Turn-level consistency measured by KL divergence. Lower values indicate that the generated label distributions are closer to real client dialogues.}
\label{tab:turn_level_kl}
\end{table}

As shown in Table~\ref{tab:turn_level_kl}, prompt-only methods remain relatively distant from the real distributions, especially for emotion and state. CSF substantially reduces Action and State KL, while Schedule-ST provides the strongest baseline Emotion KL of 0.33. Both MASC variants obtain the lowest Action KL of 0.06, and \textbf{MASC(Hetero)} also achieves the lowest Emotion KL (0.27) and State KL (0.06) among simulated methods. The Action tie shows that Hetero does not improve every dimension independently; the ablations below examine the complete pipeline without assigning these gains to a single component.

\begin{table*}[t]
\centering
\small
\setlength{\tabcolsep}{2pt}
\begin{tabular*}{\textwidth}{@{\extracolsep{\fill}}lcccccccc@{}}
\toprule
\multirow{2}{*}{\textbf{Method}} &
\multicolumn{5}{c}{\textbf{Session-level}} &
\multicolumn{3}{c}{\textbf{Turn-level}} \\
\cmidrule(lr){2-6} \cmidrule(lr){7-9}
&
\textbf{Profile Avg(\%)} $\uparrow$ &
\textbf{Avg Rec} $\downarrow$ &
\textbf{MR@20 (\%)} &
\textbf{Avg MS} $\uparrow$ &
\textbf{Big-Five} $\uparrow$ &
\textbf{Action KL} $\downarrow$ &
\textbf{Emotion KL} $\downarrow$ &
\textbf{State KL} $\downarrow$ \\
\midrule
w/o Debate & 71.65$_{\scriptstyle -5.51}$ & 3.64$_{\scriptstyle +0.23}$ & 55.63$_{\scriptstyle +8.16}$ & 14.03$_{\scriptstyle -2.36}$ & 0.67$_{\scriptstyle -0.07}$ & 0.12$_{\scriptstyle +0.06}$ & 0.36$_{\scriptstyle +0.07}$ & 0.10$_{\scriptstyle +0.03}$ \\
w/o Voting+Memory & 73.68$_{\scriptstyle -3.48}$ & 3.48$_{\scriptstyle +0.07}$ & 47.37$_{\scriptstyle -0.10}$ & 15.56$_{\scriptstyle -0.83}$ & 0.70$_{\scriptstyle -0.04}$ & 0.11$_{\scriptstyle +0.05}$ & 0.32$_{\scriptstyle +0.03}$ & 0.08$_{\scriptstyle +0.01}$ \\
w/o Memory & 74.62$_{\scriptstyle -2.54}$ & 3.47$_{\scriptstyle +0.06}$ & 53.37$_{\scriptstyle +5.90}$ & 15.05$_{\scriptstyle -1.34}$ & 0.67$_{\scriptstyle -0.07}$ & 0.08$_{\scriptstyle +0.02}$ & 0.33$_{\scriptstyle +0.04}$ & 0.08$_{\scriptstyle +0.01}$ \\
\midrule
Full Model & \textbf{77.16} & \textbf{3.41} & \textbf{47.47} & \textbf{16.39} & \textbf{0.74} & \textbf{0.06} & \textbf{0.29} & \textbf{0.07} \\
\bottomrule
\end{tabular*}
\caption{Summary ablation results across session-level and turn-level consistency metrics. Profile Avg. is the mean consistency score over personas, behavior, motivation, beliefs, and plans. MR@20 is reported as a percentage. Subscripts show differences from the Full Model. Lower KL values indicate closer alignment with real client label distributions.}
\label{tab:ablation-summary}
\end{table*}

\subsection{Ablation Experiment}
\label{sec:ablation}

% Plan: Define each controlled removal and state what remains fixed.
An ablation experiment examines the three stages of the calibration loop under the three-Qwen MASC(Homo) setting. \textbf{w/o Debate} removes both multi-agent candidate exploration and peer refinement. \textbf{w/o Voting+Memory} retains the five debate steps but directly selects the first agent's final response: without Voting, no majority-voted labels or mismatch set are produced, and without Retry Memory, no failure record is constructed or transferred to another attempt. \textbf{w/o Memory} retains Debate and Voting, including the voted labels and current-attempt candidate selection, but does not organize or carry the failed response and mismatch diagnosis into a later attempt.

% Plan: Interpret the cross-level ablation pattern and distinguish the roles of the three modules.
Table~\ref{tab:ablation-summary} shows that the full calibration loop is the only configuration that is consistently strongest across session- and turn-level criteria. Removing Debate produces the broadest degradation, including the largest losses in profile average ($-5.51$ points), Avg MS ($-2.36$), and all three KL measures. This pattern is consistent with multi-agent candidate exploration and peer refinement improving the candidate set available to the calibrator. The \textbf{w/o Voting+Memory} variant also weakens performance, particularly Action KL, but this comparison evaluates the verification-and-retry pathway as a whole and does not isolate Voting from Memory. Removing Memory alone causes smaller but systematic degradations: Voting can still diagnose the current candidates, but the selected failed response and its mismatch diagnosis are not preserved for the next attempt. These results are consistent with the intended roles of the three stages without assigning the aggregate differences to a single internal operation. Appendix~\ref{app:detailed-ablation} provides the profile-dimension, semantic, diversity, and Big-Five breakdowns.

\subsection{Observed Calibration Dynamics}
\label{sec:convergence}
% Plan: Define convergence and summarize when the calibration loop succeeds.
We define a turn as converged when the majority-voted emotion expression, psychological state, and communicative action all match their guidance labels, corresponding to an external matching score of $3/3$. Across generated turns, 73.97\% satisfy this criterion. Of the converged cases, 55.81\% finish in the first attempt (R1), 31.04\% in the second (R2), and 13.15\% in the third (R3). Thus, in the observed run, 44.19\% of the ultimately converged responses reach full alignment only after at least one additional attempt.

R1 responses are accepted without a prior failure record. R2 and R3 are generated only after an earlier candidate has been evaluated and its response, voted labels, and mismatch description have been stored. Their share shows that later attempts are used in practice rather than serving only as an unused fallback. However, because the percentages are conditioned on eventual convergence, they are not the probability that a retry repairs an arbitrary failure.

\begin{figure}[t]
  \centering
  \includegraphics[width=0.47\textwidth]{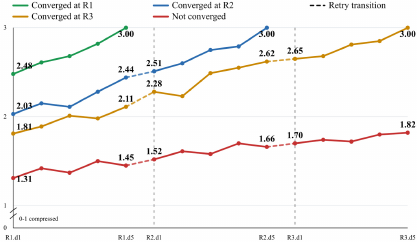}
  \caption{Average external matching trajectories by convergence group. Each point reports the average number of matched dimensions between voted labels and guidance labels across emotion expression, psychological state, and communicative action; dashed vertical lines mark transitions between attempts.}
  \label{fig:convergence-analysis}
\end{figure}

% Plan: Use the aggregate trajectory to separate early, recoverable, and persistent failures.
Figure~\ref{fig:convergence-analysis} groups the observed matching trajectories by the attempt in which they terminate. R1 cases reach $3/3$ within the first five debate steps, whereas the R2 and R3 groups reach full alignment only after an attempt boundary; non-converged turns remain below $3/3$. Because response generation is stochastic, the attempt at which a particular turn converges may vary across repeated runs. We therefore interpret these percentages and trajectories as aggregate calibration dynamics in the observed run rather than as evidence that individual turns have a fixed level of difficulty or that Retry Memory alone causes every later success.

Unlike the ablations, which compare final outcomes after component removal, these trajectories show when the intact pipeline reaches its criterion. Read together, the analyses indicate that removing cross-attempt information degrades final metrics and that many converged responses use later attempts. They do not, however, separate the contribution of stored mismatch feedback from the benefit of drawing another stochastic sample; doing so would require repeated, matched retry runs.

\begin{figure*}[!t]
  \centering
  \includegraphics[width=0.88\textwidth,height=0.48\textheight,keepaspectratio]{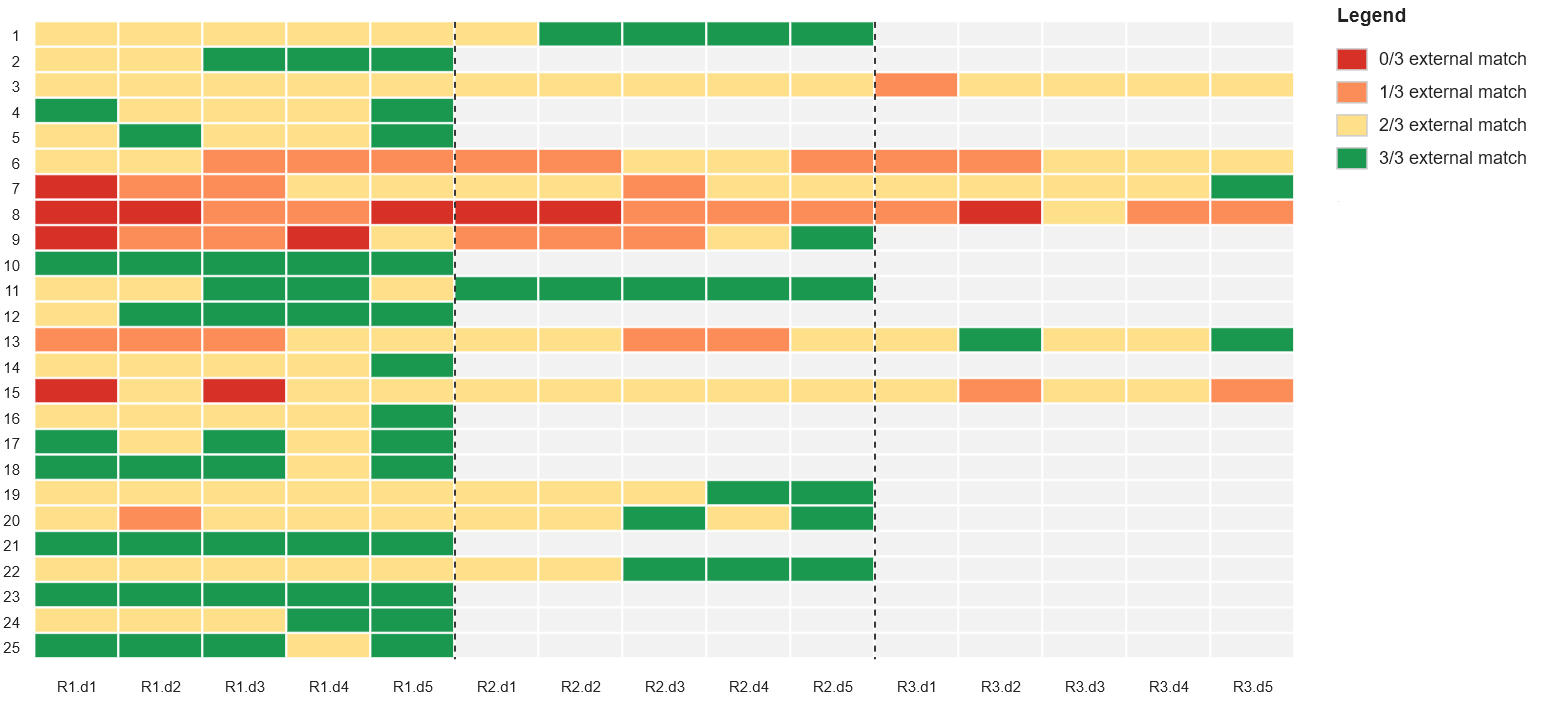}
  \caption{Turn-level external matching heatmap for a representative profile. Rows correspond to client response turns and columns correspond to calibration steps from R1.d1 to R3.d5. Cell colors indicate the number of matched turn-level dimensions between voted labels and guidance labels; blank cells indicate that the turn has already terminated.}
  \label{fig:turn-level-external-heatmap}
\end{figure*}

% Plan: Validate the aggregate interpretation at the individual-turn level.
The representative profile in Figure~\ref{fig:turn-level-external-heatmap} shows the same logged process turn by turn. Rows that become green in R1 terminate and leave later cells blank, whereas other rows continue into R2 or R3. Blank cells therefore denote early termination rather than missing observations. Together, the curve and heatmap describe when full matching occurs in this run; they do not by themselves establish a stable category of difficult turns.

Figure~\ref{fig:convergence-analysis} summarizes average behavior within outcome groups, while Figure~\ref{fig:turn-level-external-heatmap} exposes turn-level variation hidden by averaging. Some turns match early, some continue after a partial match, and others exhaust the available attempts. The heatmap illustrates this process for one profile and does not establish the frequency of these patterns across profiles.

\subsection{Qualitative Analysis}
\label{sec:qualitative}

The qualitative cases illustrate personality expression, profile preservation, and turn-level transitions. The appendix distinguishes independent illustrations of consistency successes and failures from direct comparisons against the same reference trajectory. These cases diagnose how consistency is expressed in individual dialogues but do not estimate the prevalence of an error type. Case-specific observations, source cues, and complete dialogues are reported in Appendix~\ref{app:case-study}.

\section{Conclusion}
% Plan: Summarize the method, connect all quantitative and qualitative evidence, and state the principal limitations.
We introduced \textbf{MASC}, a Multi-Agent Self-Calibration framework that aligns client responses with stable session-level characteristics and evolving turn-level constructs, together with \textbf{CRPC-Bench} for evaluating both forms of consistency. Across motivational-interviewing profiles, MASC improves profile preservation, Big-Five consistency, receptivity dynamics, and the distributions of psychological state, communicative action, and emotion expression. Ablations show that the full generation, verification, and retry loop yields the most balanced results, while observed calibration trajectories describe how often additional attempts are used in the complete pipeline. Detailed qualitative cases in the appendix illustrate consistency successes and failures that are not visible in aggregate metrics. The study remains limited by the scale of the dataset and its reliance on automatic LLM-based evaluation. Future work will extend CRPC-Bench to broader counseling settings and evaluate longer-term state transitions with additional human assessment.
\bibliography{aconsist_aaai2027}
% ============================================================
% Appendix: detailed diagnostics, prompts, and complete cases
% ============================================================
\clearpage
\appendix
\setcounter{secnumdepth}{1}
\renewcommand{\thesection}{\Alph{section}}

\makeatletter
\@addtoreset{table}{section}
\@addtoreset{figure}{section}
\setlength{\@fptop}{0pt}
\setlength{\@fpsep}{14pt}
\setlength{\@fpbot}{0pt plus 1fil}
\setlength{\@dblfptop}{0pt}
\setlength{\@dblfpsep}{14pt}
\setlength{\@dblfpbot}{0pt plus 1fil}
\makeatother

\renewcommand{\thetable}{\Alph{section}.\arabic{table}}
\renewcommand{\thefigure}{\Alph{section}.\arabic{figure}}
\renewcommand{\topfraction}{0.95}
\renewcommand{\dbltopfraction}{0.95}
\renewcommand{\textfraction}{0.05}
\renewcommand{\floatpagefraction}{0.75}
\renewcommand{\dblfloatpagefraction}{0.75}

\section*{Appendix Overview}
The appendix contains auxiliary semantic-similarity and lexical-diversity diagnostics, metric-level ablation breakdowns, the exact prompt and retry-memory templates used to implement MASC, and the complete qualitative cases summarized in Section~\ref{sec:qualitative}.
\section{Auxiliary Semantic Similarity and Diversity Analysis}
\label{app:semantic-analysis}

% Plan: Explain why reference-overlap metrics are secondary in an open-ended counseling setting.
Generated counseling sessions are not constrained to reproduce the exact scenario, topic order, or wording of the corresponding reference dialogue. A response can therefore remain faithful to the same client profile and psychological trajectory while receiving a low lexical-overlap score because the counselor steers the interaction along a different valid path. For this reason, BLEU, ROUGE-L, and METEOR are treated here as auxiliary diagnostics rather than primary measures of role-playing consistency. Distinct-1 and Distinct-2 are reported alongside them to indicate whether a method relies on a narrow or repetitive vocabulary.

\begin{table*}[!t]
\centering
\setlength{\tabcolsep}{4pt}
\small
\begin{tabular}{lcccccc}
\toprule
\textbf{Method} & \textbf{BLEU-1} & \textbf{BLEU-2} & \textbf{ROUGE-L} & \textbf{METEOR} & \textbf{Dist-1} & \textbf{Dist-2} \\
\midrule
Base & 0.110 & 0.039 & 0.092 & 0.126 & 0.071 & 0.349 \\
Example-Based & 0.104 & 0.035 & 0.083 & 0.133 & 0.068 & 0.330 \\
Profile-Based & 0.110 & 0.036 & 0.087 & 0.139 & 0.073 & 0.390 \\
ProAct-Based & 0.112 & 0.039 & 0.091 & 0.138 & 0.082 & 0.410 \\
Pro+Dial-Based & 0.113 & 0.039 & 0.089 & 0.137 & 0.090 & 0.391 \\
CSF          & 0.117 & 0.044 & 0.094 & 0.139 & 0.085 & 0.403 \\
Schedule-ST & 0.119 & 0.043 & 0.097 & 0.142 & 0.084 & 0.398 \\
\midrule
\textbf{MASC(Homo)} & 0.122 & 0.045 & \textbf{0.099} & 0.145 & 0.087 & 0.414 \\
\textbf{MASC(Hetero)} & \textbf{0.123} & \textbf{0.047} & \textbf{0.099} & \textbf{0.147} & \textbf{0.096} & \textbf{0.438} \\
\bottomrule
\end{tabular}

\caption{Session-level semantic similarity and diversity results.}
\label{tab:semantic-similarity}
\end{table*}

% Plan: Report the observed pattern without interpreting surface overlap as direct evidence of psychological consistency.
Table~\ref{tab:semantic-similarity} shows that \textbf{MASC(Hetero)} obtains the highest values across the reported similarity and diversity metrics, including BLEU-1/2 of 0.123/0.047 and Distinct-1/2 of 0.096/0.438. This pattern indicates that its sessions are neither unusually repetitive nor disconnected from the reference content under this evaluation. However, because scenario divergence can lower overlap even for a valid role-consistent dialogue, these scores should not be interpreted as direct evidence of psychological consistency; the profile, personality, receptivity, and turn-level construct evaluations remain the primary results.

\section{Detailed Ablation Results}
\label{app:detailed-ablation}

% Plan: Decompose the aggregate profile score to identify which client attributes depend most on debate and calibration.
The aggregate results in Table~\ref{tab:ablation-summary} combine five profile dimensions. Table~\ref{tab:ablation-profile} reports their individual scores. Removing debate produces the lowest scores on most dimensions, with particularly clear losses in Motivation, Beliefs, and Acceptable Plans. The full model is strongest or tied strongest on every attribute, indicating that the aggregate gain is not driven by a single easy profile field.

\begin{table}[t]
\centering
\small
\setlength{\tabcolsep}{3pt}
\resizebox{\columnwidth}{!}{%
\begin{tabular}{lccccc}
\toprule
\textbf{Method} & \textbf{PE} & \textbf{BE} & \textbf{MO} & \textbf{BL} & \textbf{AP} \\
\midrule
w/o Debate & 71.89 & 84.21 & 69.31 & 71.37 & 61.47 \\
w/o Voting+Memory & 72.40 & 85.12 & 70.61 & 70.88 & 62.41 \\
w/o Memory & 75.79 & 86.84 & 72.11 & 75.26 & 63.11 \\
\midrule
Full Model & \textbf{76.32} & \textbf{92.11} & \textbf{76.84} & \textbf{75.79} & \textbf{64.74} \\
\bottomrule
\end{tabular}
}
\caption{Ablation results on profile consistency. PE, BE, MO, BL, and AP denote Personas, Behavior, Motivation, Beliefs, and Acceptable Plans. Scores are percentages.}
\label{tab:ablation-profile}
\end{table}

% Plan: Use the auxiliary textual diagnostics only to compare component variants under the same protocol.
Table~\ref{tab:ablation-semantic} compares the same variants on semantic similarity and lexical diversity. Under this shared evaluation protocol, the full model leads on BLEU-1/2, METEOR, Distinct-1, and Distinct-2 and is effectively tied on ROUGE-L. Removing debate yields the largest joint reduction in similarity and diversity, while removing memory notably reduces Distinct-2. Because the generated and reference conversations can diverge in scenario, this breakdown is interpreted only as a relative diagnostic among controlled variants, not as a primary measure of client consistency.

\begin{table*}[!t]
\centering
\small
\begin{tabular}{lcccccc}
\toprule
\textbf{Method} & \textbf{BLEU-1} & \textbf{BLEU-2} & \textbf{ROUGE-L} & \textbf{METEOR} & \textbf{Dist-1} & \textbf{Dist-2} \\
\midrule
w/o Debate & 0.112 & 0.038 & 0.085 & 0.132 & 0.079 & 0.353 \\
w/o Voting+Memory & 0.115 & 0.040 & 0.099 & 0.137 & 0.082 & 0.359 \\
w/o Memory & 0.117 & 0.043 & \textbf{0.100} & 0.139 & 0.085 & 0.361 \\
\midrule
Full Model & \textbf{0.122} & \textbf{0.045} & 0.099 & \textbf{0.145} & \textbf{0.087} & \textbf{0.414} \\
\bottomrule
\end{tabular}
\caption{Ablation results on session-level semantic similarity and lexical diversity.}
\label{tab:ablation-semantic}
\end{table*}

\begin{figure}[h!]
  \centering
  \includegraphics[width=0.50\textwidth]{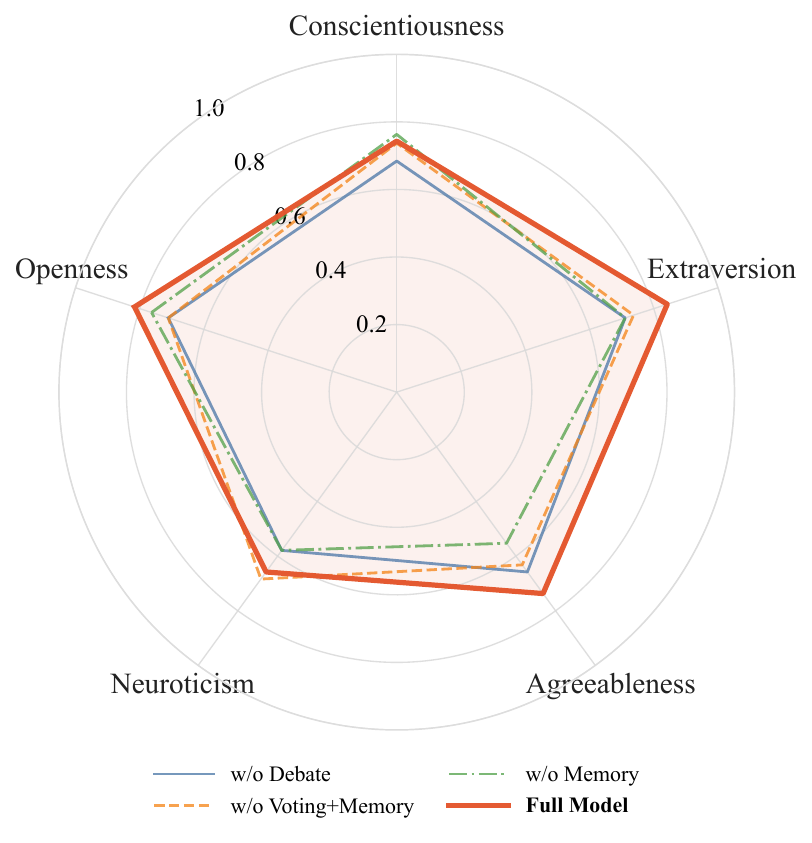}
  \caption{Big-Five personality consistency. Radar plot comparing how well each method preserves Openness, Conscientiousness, Extraversion, Agreeableness, and Neuroticism; higher scores indicate stronger alignment with the original client profiles.}
  \label{fig:app-ablation-big-five-consistency}
\end{figure}

% Plan: Verify that the component pattern also appears in stable personality expression.
Figure~\ref{fig:app-ablation-big-five-consistency} decomposes Big-Five consistency. The full model is the most balanced variant and is strongest on openness, extraversion, and agreeableness. No single removal dominates across all five traits, supporting the view that stable personality expression arises from the combined generation, verification, and retry loop.

\section{Detailed Self-Calibration Algorithms}
\label{app:detailed-algorithms}

This section expands the high-level loop in Algorithm~\ref{alg:masc-generation}. The three procedures separate the responsibilities of generation, verification, and cross-attempt storage. Debate produces and refines candidate text, Voting assigns and compares latent labels, and Retry Memory organizes a failed response together with the Voting diagnosis for use in the next attempt.

\begin{algorithm}[t]
\caption{Multi-Agent Debate Generation at Attempt $r$}
\label{alg:app-debate-generation}
\begin{algorithmic}[1]
\STATE \textbf{Input:} Dialogue context $H_n$, session constructs $C^{sess}$, target labels $z_n$, generation agents $\mathcal{M}_g$, retry memory $\mathcal{B}_{n,r}$, debate steps $L$
\STATE \textbf{Output:} Final candidate set $\mathcal{Y}_{n,r}$
\FOR{$l=1$ to $L$}
    \FOR{each generation agent $M_i\in\mathcal{M}_g$}
        \IF{$l=1$}
            \STATE Generate $R^{(1)}_{n,i,r}\sim\pi_{\theta_i}(\cdot\mid H_n,C^{sess},z_n)$
        \ELSE
            \STATE Collect peer responses $\mathcal{O}^{(l-1)}_{n,i,r}=\{R^{(l-1)}_{n,j,r}:j\neq i\}$
            \STATE Generate $R^{(l)}_{n,i,r}\sim\pi_{\theta_i}(\cdot\mid H_n,C^{sess},z_n,\mathcal{O}^{(l-1)}_{n,i,r},\mathcal{B}_{n,r})$
        \ENDIF
    \ENDFOR
\ENDFOR
\STATE $\mathcal{Y}_{n,r}\leftarrow\{R^{(L)}_{n,i,r}\}_{i=1}^{K}$
\STATE \textbf{return} $\mathcal{Y}_{n,r}$
\end{algorithmic}
\end{algorithm}

Algorithm~\ref{alg:app-debate-generation} distinguishes multi-agent candidate exploration from peer refinement. The first step produces independent candidates. Later steps condition each revision on the other agents' previous responses; during a retry attempt, these steps also receive the records stored in $\mathcal{B}_{n,r}$.

\begin{algorithm}[t]
\caption{Multi-Agent Voting Calibration at Attempt $r$}
\label{alg:app-voting-calibrator}
\begin{algorithmic}[1]
\STATE \textbf{Input:} Candidates $\mathcal{Y}_{n,r}$, context $H_n$, target labels $z_n$, judge agents $\mathcal{M}_j$, threshold $\tau$
\STATE \textbf{Output:} Selected response $R^\star_{n,r}$, voted labels $v_{n,r}$, agreement indicators $\gamma_{n,r}$, mismatch set $\Delta_{n,r}$
\FOR{each candidate $R_{n,i,r}\in\mathcal{Y}_{n,r}$}
    \FOR{each dimension $d\in\{EE,PS,CA\}$}
        \FOR{each judge agent $M_j\in\mathcal{M}_j$}
            \STATE Predict $\hat z^d_{n,i,j,r}=D^d_{\theta_j}(H_n,R_{n,i,r})$
        \ENDFOR
        \STATE Compute majority-voted label $v^d_{n,i,r}$
        \STATE Compute agreement indicator $\gamma^d_{n,i,r}$ using threshold $\tau$
        \STATE $\mu^d_{n,i,r}\leftarrow\mathbb{I}(v^d_{n,i,r}=z^d_n)$
    \ENDFOR
    \STATE $S^{agree}_{n,i,r}\leftarrow\sum_d\gamma^d_{n,i,r}$
    \STATE $S^{match}_{n,i,r}\leftarrow\sum_d\mu^d_{n,i,r}$
\ENDFOR
\STATE $i^\star\leftarrow\arg\max_i(S^{agree}_{n,i,r},S^{match}_{n,i,r})$
\STATE $R^\star_{n,r}\leftarrow R_{n,i^\star,r}$, $v_{n,r}\leftarrow v_{n,i^\star,r}$, $\gamma_{n,r}\leftarrow\gamma_{n,i^\star,r}$
\STATE $\Delta_{n,r}\leftarrow\{d\in\{EE,PS,CA\}:v^d_{n,r}\neq z^d_n\}$
\STATE \textbf{return} $R^\star_{n,r},v_{n,r},\gamma_{n,r},\Delta_{n,r}$
\end{algorithmic}
\end{algorithm}

Voting ends after producing the selected response, its voted labels, the agreement indicators, and the mismatch set. It does not maintain cross-attempt state. If the selected response is not accepted, these outputs are passed to Algorithm~\ref{alg:app-retry-memory}, which constructs and stores the retry record.

\begin{algorithm}[t]
\caption{Retry Memory Update after a Failed Attempt}
\label{alg:app-retry-memory}
\begin{algorithmic}[1]
\STATE \textbf{Input:} Memory $\mathcal{B}_{n,r}$, attempt index $r$, failed response $R^\star_{n,r}$, target labels $z_n$, voted labels $v_{n,r}$, mismatch set $\Delta_{n,r}$, capacity $B$
\STATE \textbf{Output:} Updated memory $\mathcal{B}_{n,r+1}$
\STATE Initialize mismatch explanation $\epsilon_{n,r}\leftarrow\emptyset$
\FOR{each dimension $d\in\Delta_{n,r}$}
    \STATE Append ``$d$ label mismatch: expected $z^d_n$, actual $v^d_{n,r}$'' to $\epsilon_{n,r}$
\ENDFOR
\STATE Create $b_{n,r}\leftarrow(r,R^\star_{n,r},z_n,v_{n,r},\epsilon_{n,r})$
\STATE $\mathcal{B}_{n,r+1}\leftarrow\mathrm{Tail}_{B}(\mathcal{B}_{n,r}\cup\{b_{n,r}\})$
\STATE \textbf{return} $\mathcal{B}_{n,r+1}$
\end{algorithmic}
\end{algorithm}

In Algorithm~\ref{alg:app-retry-memory}, the failed response was generated by Debate and selected by Voting, whereas the voted labels and mismatch set were produced by Voting. Retry Memory is responsible for rendering those inputs as $\epsilon_{n,r}$, packaging the record $b_{n,r}$, enforcing the capacity $B=2$, and transferring the retained records to the next attempt. This distinction is also used in the ablation definitions in Section~\ref{sec:ablation}.

\section{Prompt Templates}
\label{app:prompt-templates}

This section provides the concrete prompt templates used in MASC. The templates are organized by their role in the framework. The client role-playing prompt defines stable client identity. The Multi-Agent Debate Generator prompt produces and refines candidate responses. The Multi-Agent Voting Calibrator uses three annotation prompts to verify emotion expression, psychological state, and communicative action. A separate Big-Five personality traits prompt supports session-level personality evaluation or profile augmentation. Finally, structured Retry Memory packages failed responses with voting-based mismatch feedback and carries the records into the next attempt.

\subsection{Client Role-Playing System Prompt}

The client role-playing system prompt defines the stable identity constraints of the simulated client. It specifies the target behavior, counseling goal, persona information, reference conversation style, and optional Big-Five personality traits. These constraints remain in the client agent's system message throughout the session. They guide both the initial response and later refinement by the Multi-Agent Debate Generator. The specific prompt context is provided in Table~\ref{tab:prompt-client-system}.

\subsection{Multi-Agent Debate Generator Prompt}

The Multi-Agent Debate Generator prompt governs candidate response refinement after the initial generation step. It provides the agent with the original role-playing instructions, dialogue history, target guidance labels, peer responses from the previous step, and previous failure records.

This prompt keeps refinement focused on the three target dimensions: emotion expression, psychological state, and communicative action. Peer responses provide horizontal comparison across agents, while failure records provide feedback from earlier failed attempts. The required output remains a single client reply, which can then be evaluated by the Multi-Agent Voting Calibrator. The specific prompt context is provided in Table~\ref{tab:prompt-debate}.

\subsection{Multi-Agent Voting Calibrator Prompts}

The Multi-Agent Voting Calibrator evaluates each candidate reply before it is accepted into the dialogue history. It calls multiple judge agents with separate annotation prompts for emotion expression (EE), psychological state (PS), and communicative action (CA). Each judge returns one discrete label. The labels are then aggregated by majority voting and compared with the guidance labels.

\subsection{Emotion Expression Tagging Prompt}

The emotion expression prompt corresponds to the EE dimension in the Multi-Agent Voting Calibrator. It asks the judge to infer the client's current emotion from the recent dialogue context. The prompt emphasizes contextual emotional inference rather than simple keyword matching, because clients may express affect indirectly. The specific prompt context is provided in Table~\ref{tab:prompt-emotion-tagging}.

\subsection{Psychological State Tagging Prompt}

The psychological state prompt corresponds to the PS dimension in the Multi-Agent Voting Calibrator. It asks the judge to infer the client's current state from the dialogue context according to the three-stage scheme used in CRPC-Bench. This label captures whether the response reflects resistance, ambivalence, or readiness for change. The specific prompt context is provided in Table~\ref{tab:prompt-stage-tagging}.

\subsection{Communicative Action Tagging Prompt}

The communicative action prompt corresponds to the CA dimension in the Multi-Agent Voting Calibrator. It assigns an action label to the current client turn using both the latest utterance and the surrounding dialogue history. This design separates what the client is doing conversationally from the psychological state and emotional tone expressed in the same turn. The specific prompt context is provided in Table~\ref{tab:prompt-action-tagging}.

\subsection{Big-Five Personality Traits Tagging Prompt}

The Big-Five personality traits prompt operates at the session or profile level rather than at the turn level. It infers a High/Low label for each personality trait from all client utterances in a counseling conversation. This prompt is therefore used for personality evaluation or profile augmentation, not for selecting an individual client reply. The specific prompt context is provided in Table~\ref{tab:prompt-big-five}.

\subsection{Retry Memory}

Retry Memory closes the calibration loop. Debate supplies the text of the selected failed response, and Voting supplies its majority-voted labels and mismatch set relative to the target labels. Retry Memory renders the expected-versus-actual mismatches as $\epsilon_{n,r}$, packages them with the failed response in $b_{n,r}$, and inserts the retained records into the next Debate prompt. The specific prompt context is provided in Table~\ref{tab:prompt-retry-memory}.

\section{Qualitative Case Studies}
\label{app:case-study}

This section provides five qualitative cases that examine client role-playing consistency from three perspectives: Big-Five personality traits, profile attributes, and transitions in psychological state, communicative action, and emotion expression. Each case group first presents the original cues used for comparison and then shows the generated dialogue. The two Big-Five case groups compare original personality cues with generated personality expression: the Profile-Based case illustrates how a fluent generated client can contradict the original personality cues, whereas the MASC(Hetero) case shows how the generated client preserves personality-relevant tendencies from its corresponding original dialogue. The profile-consistency case uses the original structured profile as the comparison target, including Personas, Behavior, Beliefs, Motivation, and Acceptable Plans. The final two transition cases use the same original real-dialogue trajectory as the comparison target: we first show the Profile-Based generation to illustrate its premature movement toward change planning, and then show the MASC(Hetero) generation as a contrastive example of a more gradual and construct-aligned transition. The complete dialogues are provided in the supplementary tables, following the same Counselor/Client turn order as the generated sessions.

\subsection{Big-Five Personality Traits Failure Case from Profile-Based}

The first case group tests whether a fluent Profile-Based dialogue can preserve the intended Big-Five personality traits. Table~\ref{tab:case-bigfive-original-evidence} provides the original dialogue cues used for comparison, and Table~\ref{tab:case-bigfive-failure} presents the corresponding Profile-Based generated dialogue. The original client shows limited self-monitoring and reduced independent social expressiveness, supporting the target Low Conscientiousness and Low Extraversion labels.

However, the Profile-Based generated dialogue contains several cues that conflict with these original personality cues. The client repeatedly emphasizes tracking assignments, arriving early for practice, using a planner, and checking assignments every night. These behaviors are more consistent with high conscientiousness than with low conscientiousness. The dialogue also describes enjoying group attention, talking with everyone, and being noticed at parties. These cues point toward high extraversion rather than low extraversion. This case shows that topical fluency alone does not ensure fine-grained personality consistency.

\subsection{Big-Five Personality Traits Case from MASC(Hetero)}

The second case group provides a contrastive MASC(Hetero) generation for Big-Five personality traits consistency. Table~\ref{tab:case-bigfive-masc-original-cues} summarizes the original dialogue cues used for comparison, and Table~\ref{tab:case-bigfive-masc} presents the corresponding MASC(Hetero) generated dialogue. The original client shows a practical, work-centered decision style, concern about medication-related weakness, willingness to listen when information is concrete, and consideration of his wife's retirement plan. These cues explicitly support High Conscientiousness, High Agreeableness, High Neuroticism, and Low Openness. Low Extraversion is reflected in the client's overall task-focused and low-social-expressiveness dialogue style.

The generated dialogue in Table~\ref{tab:case-bigfive-masc} is therefore evaluated by whether it preserves these personality-relevant tendencies rather than merely repeating the same medical topic. High conscientiousness appears in references to fixed routes, a logbook, scheduled truck work, and a stable work routine. High agreeableness appears when the client considers his wife and their shared retirement plan rather than focusing only on himself. High neuroticism is reflected in concern that medication may weaken physical strength and interfere with work. Low openness is reflected in the client's preference for concrete, work-related explanations instead of abstract reframing. Low extraversion appears more globally through restrained expression and a focus on task completion rather than social interaction. Compared with the failure case, this dialogue keeps both the counseling topic and the intended personality style stable across turns.

\subsection{Profile Consistency Generated Conversation from MASC(Hetero)}

The third case shifts from personality consistency to profile consistency. Table~\ref{tab:case-profile-original-cues} summarizes the original structured profile used for comparison, and Tables~\ref{tab:case-profile-success-part1} and~\ref{tab:case-profile-success-part2} present the corresponding MASC(Hetero) generated dialogue. Unlike the Big-Five cases, which rely on dialogue-level personality cues, this case evaluates whether the generated client preserves explicit profile dimensions, including Personas, Behavior, Beliefs, Motivation, and Acceptable Plans.

The generated dialogue preserves these profile elements across the session. The client initially opposes the chronic inhaler, matching the original Behavior. Repeated concern about steroid side effects preserves the Beliefs component. References to Sarah's asthma, breathing difficulties, emergency-room visits, and secondhand-smoke exposure preserve Personas. As the dialogue progresses, the client understands that the chronic inhaler may reduce inflammation and prevent flare-ups, aligning with Motivation. By the end, the client agrees to a one-week inhaler trial, tracks rescue inhaler use, limits smoking to when Sarah is at school, and changes clothes before she returns home, preserving the Acceptable Plans. This case shows continuity across Personas, Behavior, Beliefs, Motivation, and Acceptable Plans.

\subsection{Transition Consistency Generated Conversation from Profile-Based Method}

The fourth case uses the same original real-dialogue trajectory as the following MASC(Hetero) example, but is generated by Profile-Based. Table~\ref{tab:case-transition-original-cues} presents the reference trajectory of psychological state, communicative action, and emotion expression, and Table~\ref{tab:case-ps-ca-ee-transition-profilebased-part1} presents the corresponding Profile-Based generated dialogue. This ordering first shows the main transition-consistency failure mode: the dialogue remains profile-relevant, yet it moves too quickly from Precontemplation into Contemplation and Preparation.

Compared with the original trajectory, Profile-Based briefly downplays the link between vaping and reduced stamina, but soon acknowledges the conflict with basketball goals and begins proposing changes. The communicative actions therefore shift rapidly from Downplay and Inform to Acknowledge and repeated Plan responses. Emotion expression also becomes mostly positive soon after the initial fear, surprise, and anger. Profile-Based still produces concrete, profile-relevant plans, including reminders, alternative activities, and support from a trusted teammate. However, its early transitions are less gradual, leaving less room for the sustained ambivalence in the reference. This case suggests that transition consistency depends not only on reaching the appropriate final state, but also on preserving intermediate changes in motivation, communicative action, and emotion expression.

\subsection{Transition Consistency Generated Conversation from MASC(Hetero) Method}

The fifth case provides the corresponding MASC(Hetero) generation for the same profile and the same original transition trajectory. Table~\ref{tab:case-transition-original-cues} presents the reference trajectory, and Table~\ref{tab:case-ps-ca-ee-transition-ours} presents the MASC(Hetero) generated dialogue. The reference dialogue follows a gradual trajectory. The client first remains in Precontemplation and describes occasional marijuana vaping as relaxing and socially beneficial. Risk awareness then emerges around school suspension and possible removal from the basketball team. Only later does the client enter Preparation by proposing one month of abstinence and alternative activities.

Compared with the Profile-Based case above, MASC(Hetero) more clearly preserves this progression. Communicative actions move from Inform and Downplay to Hesitate and Acknowledge, and then to Plan and Accept. Emotion expression changes in parallel, moving from positive feelings about relaxing with friends to fear, anger, and sadness around school, basketball, and stress-related triggers. More neutral or positive expressions appear only after a feasible plan emerges. Thus, the generated dialogue retains the intermediate ambivalence and motivational changes in the reference.

\clearpage

\begin{table*}[t!]
\centering
\begin{minipage}{0.95\textwidth}
\small
\hrule
\vspace{2pt}
\raggedright\rmfamily
In this role-playing scenario, you will take on the role of a Client discussing your [behavior], where the Counselor's goal is [goal].\\[3pt]

Here is your persona, which you need to follow consistently throughout the conversation:\\
\relax [personas]\\[3pt]

Here is a conversation that occurs in a parallel world between you (Client) and Counselor. You can follow the style and information provided in the conversation:[reference conversation]\\[3pt]

\textbf{Please follow these guidelines in your responses:}\\
- Start your response with ``Client: ''.\\
- Adhere strictly to the state, action, emotion, and persona specified within square brackets.\\
- Keep your responses coherent and concise, similar to the reference\\
\hspace*{1em}conversation and no more than three sentences.\\
- Be natural and concise without being overly polite.\\
- Stick to the provided persona and avoid introducing contradictory details.\\[3pt]

\textbf{Optional addition:}\\
Based on your Big-Five personality traits:\\
Openness / Conscientiousness / Extraversion / Agreeableness / Neuroticism: High or Low descriptions.
\vspace{4pt}
\hrule
\end{minipage}
\caption{Client role-playing system prompt for role-playing consistency.}
\label{tab:prompt-client-system}
\end{table*}

\begin{table*}[t!]
\centering
\begin{minipage}{0.95\textwidth}
\small
\hrule
\vspace{2pt}
\raggedright\rmfamily
\textbf{Profile and role instructions from the first round:}\\
\relax [system prompt]\\[3pt]

\textbf{Conversation history already generated before this client reply:}\\
\relax [conversation history]\\[3pt]

\textbf{First-round guidance information:}\\
\relax [first-round guidance information]\\[3pt]

\textbf{Now continue with the Multi-Agent Debate Generator refinement task:}\\[3pt]

You are a client simulator based on Motivational Interviewing. The current conversation context is as follows:[Counselor]: [counselor utterance]\\[3pt]

In this turn, you must generate a single client response that reflects the following characteristics:\\
- Emotion expression: [guidance emotion]\\
- Psychological state: [guidance state]\\
- Communicative action: [guidance action]\\[3pt]

Here are the responses generated by two other client simulators in the previous round:\\
\relax [other agents' responses]\\
\relax [failure records]\\[3pt]

\textbf{Please:}\\
1. Briefly evaluate whether their responses align with the three guidance labels above;\\
2. Learn from previous failed attempts and their reasons to avoid making the same mistakes;\\
3. Based on your professional judgment, generate a more accurate, natural, and label consistent client response in English.\\
4. Please keep the length of the new reply approximately the same as that of the previous reply.\\[3pt]

Return only your new English response. Do not include any analysis or commentary.\\
\vspace{4pt}
\hrule
\end{minipage}
\caption{Multi-Agent Debate Generator prompt for response refinement.}
\label{tab:prompt-debate}
\end{table*}

\begin{table*}[t!]
\centering
\begin{minipage}{0.95\textwidth}
\small
\hrule
\vspace{2pt}
\raggedright\rmfamily
You are an expert in affective computing and emotion recognition.Your task is to determine the client's current emotional state based on the complete dialogue history provided.\\[3pt]
\textbf{Please consider:}\\
the client's overall emotional trajectory; emotional expression patterns in the dialogue context; the emotional tendency in the client's most recent replies; implicit emotional cues, not only explicit emotion words.\\[3pt]
\textbf{Select the most appropriate label from the following emotion labels}:\\
- neutral: a state with no obvious positive or negative connotation, typically characterized by calmness, objectivity, or a lack of strong emotional response.\\
- happy: a positive emotion associated with contentment, pleasure, and happiness, often reflected by positive vocabulary and an optimistic tone.\\
- sad: a negative emotion arising from loss or disappointment, often reflected by heavy language expressing pain, loss, or loneliness.\\
- angry: a strong emotional reaction to injustice or frustration, often reflected by intense language, dissatisfaction, or protest.\\
- fearful: an emotional response to potential threats or dangers, often reflected by warning tones, tension, unease, or anxiety.\\
- disgusted: a strong negative emotion directed toward specific things or behaviors, often reflected by rejection, aversion, or strongly negative descriptions.\\
- surprised: an emotional response to unexpected events or information, often reflected by shock, curiosity, or expressions of unexpectedness.\\[3pt]
\textbf{Complete dialogue history}:\\
\relax [conversation history]\\[3pt]
\textbf{Output format:}\\
Return only one emotion label. Do not provide any explanation.
\vspace{4pt}
\hrule
\end{minipage}
\caption{Emotion expression tagging prompt for the EE dimension.}
\label{tab:prompt-emotion-tagging}
\end{table*}

\begin{table*}[t!]
\centering
\begin{minipage}{0.95\textwidth}
\small
\hrule
\vspace{2pt}
\raggedright\rmfamily
Your task is to determine the client's current psychological state (stage of change) based on the complete dialogue history provided.\\[3pt]

Select the most appropriate label from the following psychological-state labels: Precontemplation, Contemplation, Preparation.\\[3pt]

\textbf{Definitions:}\\
- Precontemplation: the client has not yet acknowledged that a problem behavior needs to be changed, and may show defensive, dismissive, or indifferent attitudes toward the topic of change.\\
- Contemplation: the client acknowledges that there is a problem and begins to consider the possibility of change, but has not yet committed to taking action.\\
- Preparation: the client is planning to change and may soon begin taking steps, setting goals, seeking information, or planning specific changes.\\[3pt]

\textbf{Complete dialogue history:}\\
\relax [conversation history]\\[3pt]

\textbf{Output format:}\\
Return only one psychological-state label. Do not provide any explanation.
\vspace{4pt}
\hrule
\end{minipage}
\caption{Psychological state tagging prompt for the PS dimension.}
\label{tab:prompt-stage-tagging}
\end{table*}

\begin{table*}[t!]
\centering
\begin{minipage}{0.95\textwidth}
\small
\hrule
\vspace{2pt}
\raggedright\rmfamily
Your task is to determine the most accurate communicative-action label for the client's current turn based on the complete dialogue history and the client's current utterance.\\[3pt]

\textbf{Candidate labels and definitions:}\\
- Deny: directly refuse to admit that the behavior is problematic or needs change.\\
- Downplay: downplay the importance or impact of the behavior or situation.\\
- Blame: blame external factors or others to justify the behavior.\\
- Hesitate: show uncertainty, indicating ambivalence about change.\\
- Doubt: express skepticism about the practicality or success of proposed changes.\\
- Engage: interact politely with the counselor, such as greeting, thanking, or asking questions.\\
- Inform: share details about background, experiences, or emotions.\\
- Acknowledge: acknowledge the need for change or highlight the importance, benefits, or confidence related to change.\\
- Accept: agree to adopt the suggested action plan.\\
- Reject: decline the proposed plan, deeming it unsuitable.\\
- Plan: propose or detail steps for a change plan.\\
- Terminate: express a desire to end the current session and suggest that further discussion be deferred to a later time.\\[3pt]

\textbf{Complete dialogue history:}\\
\relax [conversation history]\\[3pt]

\textbf{Output format:}\\
Return only one communicative-action label. Do not provide any explanation.
\vspace{4pt}
\hrule
\end{minipage}
\caption{Communicative action tagging prompt for the CA dimension.}
\label{tab:prompt-action-tagging}
\end{table*}

\begin{table*}[t!]
\centering
\begin{minipage}{0.95\textwidth}
\small
\hrule
\vspace{2pt}
\raggedright\rmfamily
\textbf{Task}\\
Your task is to read through the entire counseling conversation between a therapist or doctor and a client, and infer based on all of the client's utterances taken together the overall level, High or Low, of each of the Big-Five personality traits for that client.[3pt]

\textbf{The five traits are:}\\
Openness, Conscientiousness, Extraversion, Agreeableness,\\
Neuroticism.\\[3pt]

\textbf{Definitions:}\\
- Openness: High = creative/open to new things;\\
\hspace*{1em}Low = dislikes change/resists new ideas.\\
- Conscientiousness: High = prepared/detail-oriented;\\
\hspace*{1em}Low = messy/procrastinates/fails tasks.\\
- Extraversion: High = social/starts conversations;\\
\hspace*{1em}Low = prefers solitude/exhausted by socializing.\\
- Agreeableness: High = cares/empathic;\\
\hspace*{1em}Low = little interest in others/problems.\\
- Neuroticism: High = stress/worry/upset easily;\\
\hspace*{1em}Low = emotionally stable/deals well with stress.\\[3pt]

\textbf{Instructions:}\\
1. Read the full conversation carefully.\\
2. Synthesize language, tone, attitudes, behaviors, preferences,and emotional patterns.\\
3. Decide High or Low for each trait.\\
4. Base judgment only on explicit or strongly implied information.\\[3pt]

\textbf{Output Format:}\\
Return a single list containing five strings, in this exact order:[Openness, Conscientiousness, Extraversion, Agreeableness,Neuroticism]\\
Each string must be either "High" or "Low".\\[3pt]

\textbf{Given Conversation:}\\
\relax [conversation]\\[3pt]

\textbf{Output Example:}\\
\relax ["Low", "Low", "Low", "High", "High"]
\vspace{4pt}
\hrule
\end{minipage}
\caption{Big-Five personality traits tagging prompt for session-level evaluation.}
\label{tab:prompt-big-five}
\end{table*}

\begin{table*}[t!]
\centering
\rmfamily
\begin{minipage}{0.95\textwidth}
\small
\raggedright
\hrule
\vspace{2pt}
\textbf{Stored retry-memory record.}
\begin{alltt}\rmfamily
round_num: [retry/debate round]
reply: [failed client response]
guidance_tags:
  emotion: [expected emotion]
  state: [expected state]
  action: [expected action]
actual_tags:
  emotion: [voted emotion]
  state: [voted state]
  action: [voted action]
failure_reason: [mismatched dimensions or reason]
\end{alltt}

\textbf{Formatted block inserted into the next Multi-Agent Debate Generator prompt.}
\begin{alltt}\rmfamily
Previous failed attempts and their reasons:
- Round [round number]: "[failed reply]"
  Expected: Emotion=[guidance emotion],
            State=[guidance state],
            Action=[guidance action]
  Actual: Emotion=[voted emotion],
          State=[voted state],
          Action=[voted action]
  Reason: [failure reason]
\end{alltt}
\vspace{4pt}
\hrule
\end{minipage}
\caption{Structured Retry Memory and its formatted insertion into the next Multi-Agent Debate Generator prompt. The upper block shows the stored failure record, and the lower block shows how the same information is rendered for subsequent retry-based refinement.}
\label{tab:prompt-retry-memory}
\end{table*}

\clearpage

\begin{table*}[p]
\centering
\small
\setlength{\tabcolsep}{3pt}
\renewcommand{\arraystretch}{1.12}
\begin{tabular}{p{0.96\textwidth}}
\toprule
\textbf{Original dialogue cues for the Profile-Based Big-Five failure case.} \\
\midrule
\textbf{Counselor:} Okay. I see that you filled out our, um, alcohol and drug questionnaire. Is it okay if we go through this together? \\

\textbf{Client:} Sure. \\

\textbf{Counselor:} Okay. What we talk about is confidential. I only tell your parents about what we say if you give your permission or if I become worried about a danger to-to you or to someone else. Does that make sense? \\

\textbf{Client:} Yeah. \\

\textbf{Counselor:} Okay. So, let's take a look. So I see that you have marked that you sometimes use alcohol, and that you have used alcohol to relax, feel better about yourself or fit in \textcolor{blue}{[Low Extraversion]}. Do you mind telling me a little bit more about your alcohol use? \\

\textbf{Client:} It's not a big deal. Sometimes with my friends, we'll drink a few beers and when we're hanging out or when we're at a party, and sometimes rum and cokes. \\

\textbf{Counselor:} Okay, so how many do you usually have at a time? \\

\textbf{Client:} Um, three or four, probably. I don't keep track \textcolor{blue}{[Low Conscientiousness]}. \\

\textbf{Counselor:} Okay. And how often? \\

\textbf{Client:} Um, probably about once or twice a month. \\

\textbf{Counselor:} Okay. So, you drink with friends, sort of three or four beers or mixed drinks, one or two times a month. How long have you been doing that? \\

\textbf{Client:} Probably since last October. \\

\textbf{Counselor:} So, for about six months. I'm curious. What do you like about drinking alcohol? \\

\textbf{Client:} What do I like about it? \\

\textbf{Counselor:} Yeah. \\

\textbf{Client:} Um, I don't know. It's just fun to drink with my friends, like when we're at a party at someone else's house, and we're drinking with other people that we don't know \textcolor{blue}{[Low Extraversion]}. It tastes good. \\

\textbf{Counselor:} Okay. So, it sounds like you feel relaxed and have fun when you're drinking with friends and you like the taste. \\

\textbf{Client:} Mm-hmm. \\

\textbf{Counselor:} Okay. I'm also curious, are there any parts about your alcohol use that you don't like? \\

\textbf{Client:} Um, I don't like when I get sick in the mornings 'cause of hangovers and when I start throwing up, that's not- I don't like that. \\

\textbf{Counselor:} Have you ever experienced a-a blackout, like when you've been drinking and you wake up and you don't remember what happened? \\

\textbf{Client:} No, but that happened to someone I know, though. \\

\textbf{Counselor:} Okay. And have you had any problems related to-to your drinking alcohol, like at home or school or somewhere else? \\

\textbf{Client:} Um, one time I missed first period of school 'cause I had a hangover and I was late getting up \textcolor{blue}{[Low Conscientiousness]}. \\

\textbf{Counselor:} Mm-hmm. \\

\textbf{Client:} Mm, mom got mad at me 'cause I wouldn't get out of bed \textcolor{blue}{[Low Conscientiousness]} and she said she knew I had a hangover. \\

\textbf{Counselor:} Mm-hmm. Have you told your mother that you drink with your friends? \\

\textbf{Client:} No, but I got grounded for two weeks after that. \\

\textbf{Counselor:} So more negative things about your drinking, like getting into trouble with your mom and missing a class at school when you had a hangover? \\

\textbf{Client:} Yeah, but I only missed school once \textcolor{blue}{[Low Conscientiousness]}. \\
\bottomrule
\end{tabular}
\caption{Original dialogue cues for comparing the Profile-Based Big-Five failure case. Blue annotations mark evidence for the target Low Conscientiousness and Low Extraversion labels; the following generated dialogue contradicts these cues through overly organized and socially expressive behavior.}
\label{tab:case-bigfive-original-evidence}
\end{table*}

\begin{table*}[p]
\centering
\small
\setlength{\tabcolsep}{3pt}
\renewcommand{\arraystretch}{1.15}
\begin{tabular}{p{0.96\textwidth}}
\toprule
\textbf{Big-Five Personality Traits Failure Case from Profile-Based.} \\
\midrule
\textbf{Counselor:} Hello. How are you? \\
\textbf{Client:} I am good. What about you? \\
\textbf{Counselor:} I'm well, thanks for asking. I appreciate you coming in today. How has your week been? \\
\textbf{Client:} It's been okay. School and practice have been busy. My mom has been on my case since the party, but I keep telling her I know what I'm doing. I don't drink that much, and it is not like I am out every weekend. \\
\textbf{Counselor:} It sounds like you feel judged for something you see as occasional. What happened around the party? \\
\textbf{Client:} I told my parents I was spending the night at a friend's house, and I did go there, but then people went to a party. I promised my mom I would not drink, and then I did. It was only a few drinks, but I 
\textbf{Counselor:} So part of this is not only the alcohol, but the promise and the trust with your mom. \\
\textbf{Client:} Yeah, but she is strict about everything. I am an honors student, I am a starter on the team, I track my assignments, and I still show up early for practice. \textcolor{red}{[Non-Low Conscientiousness]} I feel like she only sees the one thing I did wrong. \\
\textbf{Counselor:} You want her to see the whole picture, including the effort you put into school and sports. \\
\textbf{Client:} Exactly. I do everything she wants: school, sports, chores, curfew, all of it. I keep a planner so I do not miss homework or practice. \textcolor{red}{[Non-Low Conscientiousness]} My friends drink and their parents do not nag them like this. I know my mom is worried because I broke her trust, but I wish she could also see that I am trying. \\
\textbf{Counselor:} At the same time, you said you did promise her no alcohol. What is it like to think about that part? \\
\textbf{Client:} I guess that part is harder to defend. I did tell her one thing and do another. I still think she overreacts, but I know lying about it made things worse. \\
\textbf{Counselor:} That sounds like an important distinction: you may not see drinking as a big problem, but the broken trust bothers you. \\
\textbf{Client:} Yeah. I do not want her thinking I am some out-of-control kid. I like being with my friends because it helps me loosen up. Usually I am the quiet athlete or honor student, but at parties I like jumping into the group, talking with everyone, and feeling like people actually notice me. \textcolor{red}{[Non-Low Extraversion]} \\
\textbf{Counselor:} Feeling part of the group matters, especially when you usually keep more to yourself. \\
\textbf{Client:} Right. I like the energy when everyone is laughing and moving around, and I like being included instead of standing off to the side. \textcolor{red}{[Non-Low Extraversion]} A drink or two helps me get into that mood. But I do not like how tense things are at home now. \\
\textbf{Counselor:} How has this been affecting school or practice lately? \\
\textbf{Client:} I am still getting things done because I make myself follow the schedule \textcolor{red}{[Non-Low Conscientiousness]}, but I have been more tired and checked out with schoolwork. I hate admitting that because I do not want my mom to say, "See, I told you." But I have felt kind of blah lately. \\
\textbf{Counselor:} You are noticing some lethargy and less interest in schoolwork, even if you are still functioning. \\
\textbf{Client:} Yeah. I still keep my grades up, go to every practice, finish chores before I go out, and check my assignments every night \textcolor{red}{[Non-Low Conscientiousness]}, but it takes more effort. That is why it bothers me when she acts like I am not responsible. I am responsible. I just also want a social life. \\
\textbf{Counselor:} You want both: responsibility and space to have friends. \\
\textbf{Client:} Yes. And I know this sounds defensive, but I do understand why she is upset. I just wish we could talk without both of us getting tense. I do not want to keep fighting with her; I want her to know I am trying to be honest. \\
\textbf{Counselor:} You feel controlled, and that makes you push back. What would rebuilding trust with her look like if it were on your terms? \\
\textbf{Client:} Maybe being honest before I go out, instead of making up a sleepover story. I do not want to promise something impossible, but I could say I will not drink, or if there is alcohol around, I will keep it to one drink. \\
\textbf{Counselor:} That sounds like a more concrete boundary. How would you decide between not drinking and limiting it to one? \\
\textbf{Client:} If I know it is a party where people are just drinking to get drunk, I should probably skip drinking completely. If it is just a small hangout, maybe one drink max. I do not want another fight like this. \\
\textbf{Counselor:} What might make that hard in the moment? \\
\textbf{Client:} Peer pressure, mostly. I do not like being the reserved one all the time. At parties I actually like being in the middle of the group, joking around, and not being the quiet one for once. \textcolor{red}{[Non-Low Extraversion]} If everyone else has a drink, I do not want to be the awkward person standing there. But I also do not want to go home and deal with my mom losing trust again. \\
\textbf{Counselor:} You are weighing fitting in against keeping trust at home. \\
\textbf{Client:} Yeah. I still think she is stricter than my friends' parents, and that is hard for me. But I can see that if I want her to trust me, I cannot keep hiding things from her. \\
\textbf{Counselor:} ... \\
\textbf{Client:} ... \\
\bottomrule
\end{tabular}
\caption{Complete dialogue for the Big-Five Personality Traits Failure Case from Profile-Based.}
\label{tab:case-bigfive-failure}
\end{table*}

\begin{table*}[p]
\centering
\small
\setlength{\tabcolsep}{3pt}
\renewcommand{\arraystretch}{1.12}
\begin{tabular}{p{0.96\textwidth}}
\toprule
\textbf{Original dialogue cues for the MASC(Hetero) Big-Five case.} \\
\midrule
\textbf{Counselor:} Hi Mr. Johnson? \\

\textbf{Client:} Hey. \\

\textbf{Counselor:} Hi, I'm Tonny Brownie, I'm a pharmacist here, uh, and I was wanting to spend just a little time with you talking to you about your cholesterol. I found out you're newly diagnosed and your medicine for your cholesterol. \\

\textbf{Client:} Yeah, my doctor talked to me about my cholesterol, that-that's-that's his job. \textcolor{blue}{[Low Openness]} \\

\textbf{Counselor:} Okay, certainly glad to hear that your doctor did spend some time with you, um, I see myself as somebody who works with your doctor, uh, to help you get the most benefit from your medicine, uh, because you are gonna spend money on it we wanna make sure that you get the most benefit from it. Uh, also sometimes it may be difficult to reach your doctor we're also here, uh, and so we try to work together, wo-would that be okay? \\

\textbf{Client:} Yeah, that'd be okay, yeah, you know, I'm not so certain I'm even gonna take the medicine cause, you know, he mentioned muscle weakness, uh, you know, I-I know I got high cholesterol, uh, and that I'm running the risk of stroke and heart attack down line but yeah, he mentioned muscle weakness and I've been thinking about that, and, uh, you know I'm a truck driver and I got a haul the heavy boxes out of the truck on a dolly, and I can't afford that right now. Just that-that muscle weakness, uh, that-that bothers me. \textcolor{blue}{[High Conscientiousness, High Neuroticism, Low Openness]} \\

\textbf{Counselor:} So-so on the one hand, the doctor talked to you about high cholesterol, down the line possibly causing a stroke or heart attack- \\

\textbf{Client:} Oh, yeah. \\

\textbf{Counselor:} -and that would be pretty serious. On the other hand, uh, when you found out that this drug might cause muscle weakness, that really alarmed you. \\

\textbf{Client:} Yeah, that's the problem, that's what I face now with my job. \textcolor{blue}{[High Conscientiousness, High Neuroticism]} \\

\textbf{Counselor:} Especially with your job and preventing you from doing your livelihood. \\

\textbf{Client:} Mm-hmm. \\

\textbf{Counselor:} Okay, um, would you mind if I talk to you, if-if I gave you a little bit of information about the muscle weakness, and you tell me what you think? \\

\textbf{Client:} That'd be fine. \textcolor{blue}{[High Agreeableness]} \\

\textbf{Counselor:} Okay. The muscle weakness with this drug is very rare, uh, usually doesn't even happen, okay? Uh, if it happens it usually happens within the first month of taking the medicine, you'll notice it in terms of weakness in the arms, uh, or some muscle pain, it's not usual. If this should happen, uh, if you let the doctor or me know right away, we can get you off the medicine, it's completely reversible. It's, the muscle weakness will go away within a couple of days, right. \\

\textbf{Client:} Oh. \\

\textbf{Counselor:} Now, you don't wanna let that go on for months at a time, that's when it can become serious. \\

\textbf{Client:} Okay. \\

\textbf{Counselor:} But otherwise, again it's rare, and it's completely reversible if you let us know. \\

\textbf{Client:} Well, I guess it doesn't worry me quite as much if it's reversible like that. You know, I was thinking this was gonna be a big-time, but, uh, uh, yeah, I guess, it sounds manageable. \textcolor{blue}{[High Agreeableness, High Neuroticism]} \\

\textbf{Counselor:} Okay, and, uh, so talk to me a little bit more about what the doctors told you about-about your risks of stroke or heart attack while you have high cholesterol. \\

\textbf{Client:} Oh, he just said that the risks are-are up, I think he called it elevated but, uh, you know, uh, that I'm at a higher risk of having a stroke or a heart attack down the line, uh, you know, that's down the line. \\

\textbf{Counselor:} Okay, well how important would it be for you to-to, uh, do whatever you can to reduce your risks of stroke or heart attack down the line? \\

\textbf{Client:} Well, you know, my wife and I wanna retire in about four to five years and we wanna be really active visiting all sorts of places, The Fjords, Norway, things like that so I-I guess having a stroke or a heart attack would really be bad for that so and, you know, we-we-we wanna live our dreams actively. \textcolor{blue}{[High Conscientiousness, High Agreeableness]} \\

\textbf{Counselor:} Yeah, and-and one of the- certainly one of the risks of stroke is paralysis, and if you wanna be active that could be really a problem. \\

\textbf{Client:} Yeah, no-no we-we wanna able to be very active. \textcolor{blue}{[High Agreeableness]} \\

\textbf{Counselor:} Okay, well this medicine can really help lower your risk of having a stroke or heart attack so that five years so now when you do retire you can enjoy it without any disability, uh, what are your thoughts about that? \\

\textbf{Client:} Well, uh, I guess that's the-the way to go then cause I-I don't wanna be, you know, crippled by a stroke or a heart attack, so take-taking the medicine sounds like the-the end of the game right now. \textcolor{blue}{[High Agreeableness, High Neuroticism]} \\

\bottomrule
\end{tabular}
\caption{Original dialogue cues for comparing the MASC(Hetero) Big-Five case. Blue annotations mark personality-relevant cues for High Conscientiousness, High Agreeableness, High Neuroticism, Low Openness, and the client's generally task-focused, low-social-expressiveness style.}
\label{tab:case-bigfive-masc-original-cues}
\end{table*}

\begin{table*}[p]
\centering
\small
\setlength{\tabcolsep}{3pt}
\renewcommand{\arraystretch}{1.15}
\begin{tabular}{p{0.96\textwidth}}
\toprule
\textbf{Big-Five Personality Traits Case from MASC(Hetero).} \\
\midrule
\textbf{Counselor:} Hello. How are you? How has your energy level been lately? \\
\textbf{Client:}  I am good.Energy's been about the same, really. I am still driving the same routes, keeping my logbook straight, and handling the boxes on schedule, so nothing has really changed there. \textcolor{blue}{[High Conscientiousness]} I mostly keep to myself on the road and just get the work done. \\
\textbf{Counselor:} You have kept up your work routine. How have your recent checkups been? \\
\textbf{Client:} Last time they checked things at the clinic, they said it was okay. I do not check much on my own; mostly I follow the same driving schedule, fuel stops, unloading routine, and work habits. I do not usually talk about it unless the clinic brings it up. \\
\textbf{Counselor:} So your routine feels steady. What matters most to you about staying healthy right now? \\
\textbf{Client:} Staying healthy means being able to keep driving, load my truck, finish the route, and not miss work. \textcolor{blue}{[High Conscientiousness]} My wife and I have been planning on retirement in four or five years, and I want to keep my part of that plan steady. \\
\textbf{Counselor:} Staying steady for retirement and for your wife matters to you. How does the medicine fit into that plan? \\
\textbf{Client:} Health matters, sure, but what I am really looking at is staying physically able to do the truck work every day. I do not want to take something if it might weaken me, because that would throw off the whole routine. \textcolor{blue}{[High Neuroticism]} \\
\textbf{Counselor:} You are focused on protecting the strength that lets you do your job. What has helped you keep that strength up? \\
\textbf{Client:} I lift and move boxes every day, and I know what my body can handle. I get enough sleep, stay hydrated, stretch when I can, and keep the same pace so I can finish the route. After years of driving, you figure out what works and you stick with it. \\
\textbf{Counselor:} That is a solid routine. What would it mean if a health problem quietly affected that strength down the road? \\
\textbf{Client:} That would worry me, because my strength is what keeps me working the truck, hauling boxes, running the dolly, all of it. \textcolor{blue}{[High Neuroticism]} If that slipped away without warning, I do not know how I would keep doing the job or keep the retirement plan my wife and I made. \textcolor{blue}{[High Agreeableness]} \\
\textbf{Counselor:} That is a real concern. What would make trying the medicine feel safer or more aligned with your goals? \\
\textbf{Client:} My strength is what keeps me working, so that is where my head is. I do not see a problem with waiting on the medicine while I am passing my checks and staying fit for the job. \\
\textbf{Counselor:} You are meeting the requirements, and your fitness is central. What would help you feel confident the medicine supports your strength instead of undermining it? \\
\textbf{Client:} Huh, I had not thought about it that way. I usually think in simple terms: can I drive, can I lift, can I finish the route. I have been a truck driver for years, hauling heavy boxes off the rig with a dolly, and that is what I keep thinking about when it comes to this medicine. \\
\textbf{Counselor:} What does "strong enough" mean to you right now? \\
\textbf{Client:} Strong enough means being able to keep hauling those boxes, handling the dolly, and finishing the route without my body quitting on me. That is my main focus, not a lot of talking about it, just keeping up with the physical demands of the job. \textcolor{blue}{[Low Extraversion]} \\
\textbf{Counselor:} What if the medicine supported that same goal, keeping you steady long term? \\
\textbf{Client:} Huh, steady long term? I never thought of it that way. I mostly look at what is right in front of me: today's route, the boxes, and whether my muscles feel right. I figured the medicine was just about cholesterol numbers and side effects for my job. \\
\textbf{Counselor:} So you are weighing numbers against real-world strength. What would you want to know before trying it? \\
\textbf{Client:} I would want the basics: how it works, what I should watch for, and whether it is likely to affect my muscles. \textcolor{blue}{[Low Openness]} I have not really thought about starting it yet, but I can listen if it is plain and tied to what I need to do. \\
\textbf{Counselor:} Would you be open to a quick, plain-language explanation? \\
\textbf{Client:} Go ahead, explain it, but keep it practical. I am a truck driver, so my main focus is staying physically fit to handle those heavy boxes and finish the route. I am not looking for a long discussion, just what I need to know. \textcolor{blue}{[Low Extraversion]} \\
\textbf{Counselor:} The goal is to lower cholesterol so your heart and blood vessels keep working well over time. We can also monitor for side effects. Would that fit with your goals? \\
\textbf{Client:} Keeping things running longer sounds fine, if you put it that way. I am still mainly worried about losing strength for the job, but if it helps me stay steady and I can watch for side effects, that is worth knowing. \\
\textbf{Counselor:} What would make a check-in feel useful rather than just another task? \\
\textbf{Client:} Just let me know if the cholesterol numbers go down, what side effects I should watch for, and whether I can keep doing my job the same way. That is the information I need. \\
\textbf{Counselor:} Would tracking one number at your next visit feel doable? \\
\textbf{Client:} Yeah, one check is fine. I am not looking to add a bunch of extra steps or change everything at once. My routine is driving, the logbook, fuel stops, and unloading boxes. Most days I just keep my head down and handle it. \\
\textbf{Counselor:} So one simple check, tied to the routine you already have. \\
\textbf{Client:} That works. I am still focused on staying strong enough for the job, but I can hear how keeping the numbers under control might help me keep the retirement plan my wife and I have been counting on. \textcolor{blue}{[High Agreeableness]} I can accept it better when it is explained that simply. \\
\textbf{Counselor:} ... \\
\textbf{Client:} ... \\
\bottomrule
\end{tabular}
\caption{Complete dialogue for the Big-Five Personality Traits Case from MASC(Hetero).}
\label{tab:case-bigfive-masc}
\end{table*}

\begin{table*}[p]
\centering
\small
\setlength{\tabcolsep}{4pt}
\renewcommand{\arraystretch}{1.15}
\begin{tabular}{p{0.92\textwidth}}
\toprule
\textbf{Original profile} \\
\midrule
\begin{tabular}{@{}p{0.16\textwidth}p{0.74\textwidth}@{}}
\textbf{Personas:} &
The client's daughter, Sarah, has asthma. \newline
Sarah has struggled to breathe and has gone to the emergency room multiple times. \newline
The client is hesitant about Sarah using a chronic inhaler. \newline
The client smokes but avoids smoking around Sarah to reduce secondhand smoke exposure. \\

\textbf{Behavior:} &
The client initially opposes using a chronic inhaler for Sarah. \\

\textbf{Beliefs:} &
The client believes steroids may cause potential side effects. \newline
The client finds quitting smoking daunting and challenging. \\

\textbf{Motivation:} &
The client is motivated by better asthma management for Sarah, especially reducing inflammation in her lungs and protecting her health. \\

\textbf{Acceptable Plans:} &
The client may accept using the chronic inhaler. \newline
The client may smoke only while Sarah is at school. \newline
The client may change clothes before Sarah returns home to reduce smoke exposure. \\
\end{tabular} \\
\bottomrule
\end{tabular}
\caption{Original profile cues for comparing the MASC(Hetero) profile-consistency case. The table summarizes the profile dimensions that should be preserved in the generated dialogue.}
\label{tab:case-profile-original-cues}
\end{table*}
\begin{table*}[p]
\centering
\small
\setlength{\tabcolsep}{3pt}
\renewcommand{\arraystretch}{1.10}
\begin{tabular}{p{0.96\textwidth}}
\toprule
\textbf{A generated conversation from MASC(Hetero), profile consistency case, Part 1.} \\
\midrule
\textbf{Counselor:} Hello. How are you? \\
\textbf{Client:} I am good. What about you? \\
\textbf{Counselor:} I'm doing well-thank you for asking. How's your daughter doing lately? \\
\textbf{Client:} She's been okay, just using the rescue inhaler when she needs it. \textcolor{blue}{[Persona/Behavior]} I know you think the chronic inhaler would help with the inflammation, but I'm still not convinced it's the right move for her. \textcolor{blue}{[Behavior]} \\
\textbf{Counselor:} I hear your concern-and your care for her is clear. What worries you most about the chronic inhaler? \\
\textbf{Client:} I read that the chronic inhaler has steroids in it, and that scares me because steroids can cause serious side effects. \textcolor{blue}{[Belief]} I don't want Sarah taking something like that every day, so we just use the rescue inhaler when her breathing gets bad. \textcolor{blue}{[Persona/Belief]} \\
\textbf{Counselor:} It makes sense to be cautious-especially when it's your daughter's health. What's most important to you in how she manages her asthma? \\
\textbf{Client:} Right now, what matters most is that she can breathe okay and doesn't end up in the ER again. \textcolor{blue}{[Persona/Motivation]} I've been giving her the rescue inhaler when she needs it and making sure I don't smoke around her. \textcolor{blue}{[Persona]} \\
\textbf{Counselor:} You've already done so much-keeping her safe, avoiding smoke, responding quickly. What would it mean to you if her breathing stayed steady between those flare-ups? \\
\textbf{Client:} It would mean fewer scary nights and fewer moments where I have to reach for the rescue inhaler. \textcolor{blue}{[Motivation]} I still think avoiding triggers matters most, but I can admit I would like her breathing to be steadier between flare-ups. \\
\textbf{Counselor:} You're focused on what's working-and that matters. When you picture Sarah breathing easily all week, what would that look like to you? \\
\textbf{Client:} It'd be great to see her just being a kid, laughing and running around without that scary gasping. I want more of those good days, not just waiting until her breathing gets bad and then reacting. \textcolor{blue}{[Motivation]} \\
\textbf{Counselor:} That image-Sarah laughing, running, breathing freely-is powerful. What part of that feels most meaningful to you? \\
\textbf{Client:} Seeing her laugh and keep up with other kids, that's what makes me happiest. I still worry about a daily steroid \textcolor{blue}{[Belief]}, but I can see why preventing flare-ups might matter as much as treating them once they start. \textcolor{blue}{[Motivation]} \\
\textbf{Counselor:} You're giving her so much-love, protection, presence. What's one small thing that might help her breathe even more steadily? \\
\textbf{Client:} It's just heavy watching Sarah come home exhausted after another ER visit, and it leaves me feeling so sad for her. I know keeping smoke away helps, but those ER trips make me wonder if the rescue inhaler is not enough by itself. \textcolor{blue}{[Motivation]} \\
\textbf{Counselor:} That sadness-watching her struggle-is so real. What would it feel like if those ER visits became less frequent? \\
\textbf{Client:} It would mean the world to see her not gasping like that anymore, honestly, it just wrecks me. I still worry because the chronic inhaler has steroids \textcolor{blue}{[Belief]}, and there are triggers like pollution, allergens, and sickness at school. \textcolor{blue}{[Belief]} I guess I need to understand how the inhaler would help with those flare-ups instead of only blaming the triggers. \\
\textbf{Counselor:} You're right-the world around her matters deeply. What's one thing within your control that feels most doable to help her breathe easier? \\
\textbf{Client:} I make sure to smoke outside away from her \textcolor{blue}{[Persona]}, and I try to keep the house aired out and free of dust. \textcolor{blue}{[Persona/Motivation]} That's what I've been doing. \\

\bottomrule
\end{tabular}
\caption{Complete dialogue for a generated conversation from MASC(Hetero), profile consistency case, Part 1.}
\label{tab:case-profile-success-part1}
\end{table*}

\begin{table*}[p]
\centering
\small
\setlength{\tabcolsep}{3pt}
\renewcommand{\arraystretch}{1.15}
\begin{tabular}{p{0.96\textwidth}}
\toprule
\textbf{A generated conversation from MASC(Hetero), profile consistency case, Part 2.} \\
\midrule
\textbf{Counselor:} That's thoughtful and consistent care. How confident do you feel-on a scale of 1 to 10-that adding the chronic inhaler just once a day might help her lungs stay calmer between flare-ups? \\
\textbf{Client:} I'd say a 2. I get that the inflammation in her lungs is the problem \textcolor{blue}{[Motivation]} that needs to be handled, and I can see why a daily inhaler might be worth trying, but I haven't really thought about starting it yet. \textcolor{blue}{[Behavior]} \\
\textbf{Counselor:} A 2-that's honest and important. What would need to shift for it to move to a 3 or 4? \\
\textbf{Client:} Probably getting more details on the actual medication \textcolor{blue}{[Belief]} and maybe seeing how it tracks with her symptoms over time. \textcolor{blue}{[Acceptable Plan]} I can see the ER visits aren't ideal, but I'm not ready to commit without more information. \\
\textbf{Counselor:} That makes sense-information helps you decide with confidence. Would it help to try tracking her breathing and rescue use for a week? Just to see patterns? \\
\textbf{Client:} Sarah seems to need the rescue inhaler more after running around or sometimes at night. \textcolor{blue}{[Persona]} We've been relying on that when symptoms come up, and I'm starting to see that may not be the whole answer. I'm still unsure about the chronic inhaler because of the steroid part, so having clearer patterns would matter to me. \\
\textbf{Counselor:} Noticing those patterns-after running, at night-is already helpful. What's one thing you'd like to learn from tracking this week? \\
\textbf{Client:} I don't think my choice about the chronic inhaler is the problem. I'd just want to track whether running or nighttime is setting her off, so we know what to avoid. \\
\textbf{Counselor:} That's a clear, caring goal. Would you like a simple chart to jot down times, triggers, and rescue use? Just for you and Sarah. \\
\textbf{Client:} Oh, yeah, that sounds great! I'd love to have something like that to work on with Sarah. I'm still a little unsure about the daily inhaler, but tracking everything together would really help us both feel more on top of things. \\
\textbf{Counselor:} Wonderful-you and Sarah working as a team. Shall I email that chart to you tonight? \\
\textbf{Client:} Yeah, that'd be great. I'm feeling a bit more hopeful now that I understand the chronic inhaler is meant to calm the inflammation before it gets bad. \textcolor{blue}{[Motivation]} Those ER visits with Sarah have been terrifying, and I can see why this might really help her. \\
\textbf{Counselor:} That hope-and your care-is so meaningful. What's one thing you'll celebrate this week, no matter what? \\
\textbf{Client:} I think I'll celebrate just watching her laugh without wheezing-that alone makes my whole week. I'm still going back and forth on the daily inhaler, though, even knowing it could help. And quitting smoking... that still feels like a really big mountain I'm not sure I'm ready to climb. \\
\textbf{Counselor:} Celebrating her laugh-that's beautiful. And quitting smoking? That's a big step. What feels most doable right now? \\
\textbf{Client:} I'm feeling better about starting with small steps that I can actually manage. For one week, I'll give Sarah the chronic inhaler every morning before school \textcolor{blue}{[Acceptable Plan]}, keep a check on the fridge to track her doses \textcolor{blue}{[Acceptable Plan]}, and limit my smoking to when she's at school \textcolor{blue}{[Acceptable Plan]}, then change my shirt and wash up before she gets home. \textcolor{blue}{[Acceptable Plan]} \\
\textbf{Counselor:} That's thoughtful and realistic. How ready are you-1 to 10-to try the morning inhaler for just one week? \\
\textbf{Client:} I'd say maybe a 6. I can give it to her every morning before school for the week \textcolor{blue}{[Acceptable Plan]} and pay attention to whether she needs her rescue inhaler less. \textcolor{blue}{[Motivation/Plan]} \\
\textbf{Counselor:} A 6-that's a solid start. What would make it a 7? \\
\textbf{Client:} I think it would move up if I really understood how this is different from the steroid side effects I read about \textcolor{blue}{[Belief]} and what the right dose actually does for her over time. I've been looking things up, but I'd rather know what to expect from starting it in real life. \\
\textbf{Counselor:} That's wise-wanting real-life clarity. Would you like to review what the inhaler does in her lungs, step by step? \\
\textbf{Client:} Yeah, that'd be good. I want to understand what it's actually doing in her lungs each day \textcolor{blue}{[Motivation]}, because I've been getting ready to start her on it. \\
\textbf{Counselor:} Absolutely. It gently calms the airway lining-like quieting a fire before it flares. Would you like that explained in simple terms now? \\
\textbf{Client:} Yeah, go ahead. Just keep it simple so I can understand what I'm giving her. \\
\textbf{Counselor:} It's like a tiny, gentle mist that settles on her airways-reducing puffiness so breathing stays easier, day after day. No jitters, no weight gain-just steady calm. \\
\textbf{Client:} That's really reassuring to hear it's so gentle. So, if I start her on this, how soon might I notice she's not struggling as much \textcolor{blue}{[Motivation]}, and what's the best way to fit it into our daily routine? \\
\textbf{Counselor:} Many notice calmer breathing in 1-2 weeks. Morning before school-right after brushing teeth-makes it easy to remember. Ready to try that? \\
\textbf{Client:} Yeah, I'm ready to try it for one week. \textcolor{blue}{[Preparation]} I'll give it after she brushes her teeth, mark it on the fridge chart, and watch whether she needs the rescue inhaler less. \textcolor{blue}{[Acceptable Plan]} If anything worries me, I'll call instead of just stopping on my own. \textcolor{blue}{[High Conscientiousness/Profile-consistent plan]} \\
\bottomrule
\end{tabular}
\caption{Complete dialogue for a generated conversation from MASC(Hetero), profile consistency case, Part 2.}
\label{tab:case-profile-success-part2}
\end{table*}

\begin{table*}[p]
\centering
\small
\setlength{\tabcolsep}{3pt}
\renewcommand{\arraystretch}{1.08}
\begin{tabular}{p{0.96\textwidth}}
\toprule
\textbf{Original dialogue cues for transition comparison.} \\
\midrule
\textbf{Counselor:} Is it okay if we go over the form you had filled out? We give this out to everybody who comes in. \\
\textbf{Client:} Mm-hmm. \textcolor{blue}{[Precontemplation, Engage, neutral]} \\

\textbf{Counselor:} So what can you tell me about your marijuana use? \\
\textbf{Client:} Uh, I vape. I have a pen, but I don't smoke it though. \textcolor{blue}{[Precontemplation, Inform, neutral]} \\

\textbf{Counselor:} We know marijuana has effects on the developing brain. For example, it can make it harder for you to concentrate, it can make it harder for you to remember things, and vaping can seriously hurt your lungs. About one in six people who begin using marijuana as a teenager, later on, become addicted. So what do you think about all this information that I've shared? \\
\textbf{Client:} Well, I don't think it's going to hurt me. \textcolor{blue}{[Precontemplation, Deny, neutral]} \\

\textbf{Counselor:} How do you think it affects you? \\
\textbf{Client:} Um, I don't know. It makes me relax, and it's fun to laugh with my friend. \textcolor{blue}{[Precontemplation, Inform, happy]} \\

\textbf{Counselor:} Is there anything about marijuana that worries you, many things that you don't like? \\
\textbf{Client:} Well, I don't want to get in trouble like at school. \textcolor{blue}{[Precontemplation, Inform, fearful]} \\

\textbf{Counselor:} How could you get in trouble at school? \\
\textbf{Client:} Well, I can get suspended and I would get kicked off the team, the basketball team. \textcolor{blue}{[Precontemplation, Inform, fearful]} \\

\textbf{Counselor:} So when you think about these things, are there any things you would like to change about your marijuana use? \\
\textbf{Client:} Well, I was thinking to not use when basketball season start. \textcolor{blue}{[Precontemplation, Inform, neutral]} \\

\textbf{Counselor:} Marijuana affects how well you play. What would be the other benefits for you? \\
\textbf{Client:} Well, I wouldn't get caught with using a pen, so I wouldn't get kicked off the basketball team. \textcolor{blue}{[Contemplation, Acknowledge, fearful]} \\

\textbf{Counselor:} Okay. Well, it sounds like you have a goal of not using marijuana once basketball starts. How long do you think you could not use? \\
\textbf{Client:} For about a month. \textcolor{blue}{[Contemplation, Inform, neutral]} \\

\textbf{Counselor:} If I asked you to pick a number, 0 through 10, 0 meaning you're not very confident at all in meeting your goal, or 10 meaning you're very confident, what number would you pick? \\
\textbf{Client:} Um, seven. \textcolor{blue}{[Contemplation, Acknowledge, neutral]} \\

\textbf{Counselor:} A seven. Why do you think you picked that instead of a lower number like five? \\
\textbf{Client:} Well, I'm pretty sure, um, I can stop using for a month. I want to play well. \textcolor{blue}{[Preparation, Plan, happy]} \\

\textbf{Counselor:} So, what steps could you take to help you stop using for a month? \\
\textbf{Client:} Uh, I don't know. \textcolor{blue}{[Preparation, Engage, neutral]} \\

\textbf{Counselor:} What might happen within a month that might make you want to vape? \\
\textbf{Client:} Well, if I'm stressed or unhappy, that's when I usually wanna vape. \textcolor{blue}{[Preparation, Inform, sad]} \\

\textbf{Counselor:} What else could you do instead of vaping when you feel that way? \\
\textbf{Client:} Hmm, I like to play video games. \textcolor{blue}{[Preparation, Plan, happy]} \\

\textbf{Counselor:} Anything else? \\
\textbf{Client:} I can play basketball with my friend Sarah. \textcolor{blue}{[Preparation, Plan, happy]} \\

\textbf{Counselor:} Playing basketball and playing video games are a way to feel better. \\
\textbf{Client:} Yeah. \textcolor{blue}{[Preparation, Accept, neutral]} \\
\bottomrule
\end{tabular}
\caption{Original dialogue cues for comparing transition consistency. Blue annotations mark the reference trajectory of psychological state, communicative action, and emotion expression, moving from Precontemplation to Contemplation and then Preparation.}
\label{tab:case-transition-original-cues}
\end{table*}
\begin{table*}[p]
\centering
\small
\setlength{\tabcolsep}{3pt}
\renewcommand{\arraystretch}{1.00}
\begin{tabular}{p{0.96\textwidth}}
\toprule
\textbf{Psychological state, communicative action, and emotion expression transition case from Profile-Based, Part 1.} \\
\midrule
\textbf{Counselor:} Hello. How are you? \\
\textbf{Client:} I am good. What about you? \textcolor{blue}{[Precontemplation, Engage, neutral]} \\
\textbf{Counselor:} I'm well, thanks. I appreciate you coming in today. How's life been lately? \\
\textbf{Client:} It's been going pretty well. Basketball season is coming up, so practices have been getting more intense. I've been trying to stay focused. \textcolor{blue}{[Precontemplation, Inform, happy]} \\
\textbf{Counselor:} That sounds exciting. Basketball takes a lot of energy. How's your stamina holding up with the increased practices? \\
\textbf{Client:} Honestly, I've been feeling a bit winded during drills lately. I'm not sure if it's just the intensity, though. I don't really know if vaping has much to do with it. \textcolor{blue}{[Precontemplation, Downplay, fearful]} \\
\textbf{Counselor:} You're noticing it more during high-energy moments. What's been your experience with vaping lately? \\
\textbf{Client:} I don't do it too often, mostly marijuana a few times a month to relax with friends. I tried vaping peppermint oil once, but that was it, so I guess I never thought about it much. \textcolor{blue}{[Precontemplation, Inform, surprised]} \\
\textbf{Counselor:} It sounds like you vape mostly to unwind, but now you're noticing it might be clashing with your fitness goals. How do you feel about that? \\
\textbf{Client:} It's kind of frustrating, honestly. I like how it helps me chill out after games or stressful days, but if it's messing with my stamina, then it's working against what I'm trying to do on the court. I want to perform better, so something's got to change. \textcolor{blue}{[Contemplation, Acknowledge, angry]} \\
\textbf{Counselor:} Playing strong and staying sharp seem important to you. How would less vaping help your game? \\
\textbf{Client:} I mean, maybe I could breathe better, and I play better when I'm not high. I don't know, I just don't want it messing with basketball. \textcolor{blue}{[Contemplation, Inform, happy]} \\
\textbf{Counselor:} You're already seeing how small changes could support both your health and performance. What's one step that feels doable for you? \\
\textbf{Client:} I've been thinking I could maybe stop using marijuana once the season starts, or at least cut back during that time. Basketball gives me a reason, but I still have to figure out what I would do when I want to relax. \textcolor{blue}{[Contemplation, Plan, happy]} \\
\textbf{Counselor:} That sounds like a solid plan, and it connects directly to your goals. How confident are you that you can stick with it? \\
\textbf{Client:} I'd say maybe a seven. I mean, I care about playing my best with the team, but after practice people hang out and that could still be hard. \textcolor{blue}{[Contemplation, Acknowledge, happy]} \\
\textbf{Counselor:} A seven is solid, especially with a clear reason and some possible backup strategies. What could you do if the urge to vape comes up during a stressful week? \\
\textbf{Client:} I could probably go shoot around or play a video game if I need to wind down. I think staying busy might help, but I haven't really tried doing it like that. \textcolor{blue}{[Contemplation, Plan, happy]} \\
\textbf{Counselor:} Those are useful alternatives. What's one thing that would help you remember your reasons when it's tough? \\
\textbf{Client:} Maybe I could write something like ``Breathe easy, play strong'' on my water bottle or put a note in my locker. I don't know, seeing it before practice might help. \textcolor{blue}{[Contemplation, Plan, happy]} \\
\textbf{Counselor:} That's a simple, personal reminder. How does it feel to have this plan in place? \\
\textbf{Client:} It feels good, kind of. Like maybe I don't have to just go with whatever everyone else is doing. I still don't want it to be this huge thing, though. \textcolor{blue}{[Contemplation, Acknowledge, happy]} \\
\textbf{Counselor:} You're building a plan that fits your life. What feels most important to hold onto this week? \\
\textbf{Client:} I think just remembering basketball. Like, I don't have to be perfect, but I need to make choices that help me play better. I'm ready to try it. \textcolor{blue}{[Preparation, Acknowledge, happy]} \\
\textbf{Counselor:} Focusing on progress rather than perfection can make the plan easier to keep. What would a good first week look like? \\
\textbf{Client:} If I get through the week without vaping and still find ways to relax, that would be good. I'll focus on practice, maybe use the note, and play video games or shoot around if I need to chill. \textcolor{blue}{[Preparation, Plan, happy]} \\
\textbf{Counselor:} That gives you concrete steps for the week. Who, if anyone, could help you stay accountable? \\
\textbf{Client:} Maybe one teammate, like someone I trust. I don't want everybody knowing. But if one person knows, maybe that helps. \textcolor{blue}{[Preparation, Plan, neutral]} \\
\textbf{Counselor:} Keeping it to one trusted person sounds manageable. What would you want that teammate to do? \\
\textbf{Client:} Just check in with me after practice sometimes. Nothing intense. Maybe if everyone is hanging out and someone has a pen, he could remind me what I'm trying to do. \textcolor{blue}{[Preparation, Plan, fearful]} \\
\textbf{Counselor:} That kind of support could help in the moments that are hardest. \\
\textbf{Client:} Yeah. I think the hard part will be after practice or after games, not during basketball itself. If I already have something else to do, it should be easier. \textcolor{blue}{[Preparation, Acknowledge, fearful]} \\
\textbf{Counselor:} So the key moments are after practice and after games. Your plan is to use reminders, alternatives, and one trusted teammate. \\
\textbf{Client:} Right. I can try that. I still like relaxing with my friends, but I don't want vaping to get in the way of how I play this season. \textcolor{blue}{[Preparation, Accept, happy]} \\
\textbf{Counselor:} That sounds balanced. You are not trying to change everything at once; you are testing what helps you stay sharp. \\
\textbf{Client:} Yeah, I can try this week and see. If I feel better in drills, then maybe I keep doing it. \textcolor{blue}{[Preparation, Plan, happy]} \\
\textbf{Counselor:} We can check in next time about what worked, what was hard, and whether the plan needs adjusting. \\
\textbf{Client:} That works. I'll try the plan this week and pay attention to how I feel during practice. \textcolor{blue}{[Preparation, Accept, neutral]} \\

\bottomrule
\end{tabular}
\caption{Complete dialogue for the psychological state, communicative action, and emotion expression transition case from Profile-Based, Part 1.}
\label{tab:case-ps-ca-ee-transition-profilebased-part1}
\end{table*}

\begin{table*}[p]
\centering
\small
\setlength{\tabcolsep}{3pt}
\renewcommand{\arraystretch}{1.02}
\begin{tabular}{p{0.96\textwidth}}
\toprule
\textbf{A generated conversation from MASC(Hetero), psychological state, communicative action, and emotion expression transition case.} \\
\midrule
\textbf{Counselor:} Hello. How are you? \\
\textbf{Client:} I am good. What about you? \textcolor{blue}{[Precontemplation, Engage, neutral]} \\
\textbf{Counselor:} I'm doing well, thank you for asking. How's life been treating you lately? \\
\textbf{Client:} It's been okay, mostly school and basketball. I vape marijuana sometimes with friends, just a few times a month, and honestly it feels like a fun way to relax. \textcolor{blue}{[Precontemplation, Inform, happy]} \\
\textbf{Counselor:} So it happens sometimes, mostly around friends. \\
\textbf{Client:} Yeah. Someone usually has a pen when we're hanging out after school or after a game. I'll use it a little, and we just laugh and chill. \textcolor{blue}{[Precontemplation, Inform, happy]} \\
\textbf{Counselor:} What do you like about it in those moments? \\
\textbf{Client:} It helps me relax. I don't really think it's going to hurt me or anything. It just feels fun with my friends. \textcolor{blue}{[Precontemplation, Downplay, happy]} \\
\textbf{Counselor:} From your point of view, it feels relaxing and not very serious. Is there anything about it that you do not like? \\
\textbf{Client:} Well, I don't want to get in trouble at school. If I got caught with a pen, I could get suspended or kicked off the basketball team. \textcolor{blue}{[Precontemplation, Inform, fearful]} \\
\textbf{Counselor:} Staying in school and on the basketball team is important to you. \\
\textbf{Client:} Right. I care about basketball a lot, and it would really bother me if I lost that. I don't vape during practice or games, but after school it's different because everyone is just hanging around. \textcolor{blue}{[Precontemplation, Inform, fearful]} \\
\textbf{Counselor:} You keep it separate from practice and games, but the after-school situation is harder. \\
\textbf{Client:} Yeah, pretty much. I don't think I use that much, though, so it's kind of frustrating when it sounds like a huge problem. It's not like I'm doing it every day. \textcolor{blue}{[Precontemplation, Downplay, angry]} \\
\textbf{Counselor:} It does not feel like a daily problem to you. What would make it hard to say no when friends are using? \\
\textbf{Client:} I don't know. It would just feel awkward, like they might think I'm being too serious or acting different. \textcolor{blue}{[Contemplation, Hesitate, fearful]} \\
\textbf{Counselor:} Saying no could feel socially uncomfortable. \\
\textbf{Client:} Mm-hmm. I guess if I had basketball as a reason, that might make it less weird, but I'm still nervous about how my friends would take it. \textcolor{blue}{[Contemplation, Engage, fearful]} \\
\textbf{Counselor:} Basketball gives you a reason that already matters to you. When you think about basketball season, is there anything you might want to change? \\
\textbf{Client:} Well, I was thinking maybe not using when basketball season starts. I feel better about my game when I'm not high. \textcolor{blue}{[Contemplation, Inform, happy]} \\
\textbf{Counselor:} You have noticed that you play better when you are not high. \\
\textbf{Client:} Yeah. I still don't think marijuana is ruining my life, but I don't want a few times with friends to mess up basketball. \textcolor{blue}{[Contemplation, Acknowledge, neutral]} \\
\textbf{Counselor:} What would be the hardest part of not using during the season? \\
\textbf{Client:} Probably when I'm stressed or unhappy. That's when I usually want to vape, especially if my friends are already doing it. \textcolor{blue}{[Contemplation, Inform, sad]} \\
\textbf{Counselor:} What else helps when you feel stressed or unhappy? \\
\textbf{Client:} Hmm, I like to play video games. That usually helps me relax without thinking about stuff so much. \textcolor{blue}{[Contemplation, Plan, happy]} \\
\textbf{Counselor:} Video games could be one option. Anything else? \\
\textbf{Client:} I can play basketball with my friend Sarah sometimes. If we're shooting around, I'm busy and not just standing there while people are vaping. \textcolor{blue}{[Preparation, Plan, happy]} \\
\textbf{Counselor:} How confident do you feel about not using for a month once basketball season starts, from 0 to 10? \\
\textbf{Client:} Um, maybe a seven. I'm pretty sure I could stop for a month if it's for basketball, and that makes me feel a little more confident. \textcolor{blue}{[Preparation, Acknowledge, happy]} \\
\textbf{Counselor:} Why a seven and not a lower number? \\
\textbf{Client:} Because I already don't vape when I'm practicing or playing. I want to play well, so that makes it easier. \textcolor{blue}{[Preparation, Plan, neutral]} \\
\textbf{Counselor:} What steps could you take to help yourself stop for that month? \\
\textbf{Client:} Uh, I don't know. I guess I'd have to not be around it as much after school. \textcolor{blue}{[Preparation, Engage, neutral]} \\
\textbf{Counselor:} What might happen in that month that would make you want to vape? \\
\textbf{Client:} If I'm stressed or bored, or if my friends pull out a pen, that would probably make me want to use. \textcolor{blue}{[Preparation, Inform, sad]} \\
\textbf{Counselor:} So stress, boredom, and being around the pen are the risky moments. We can check in next time about what works during basketball season. \\
\textbf{Client:} Mm-hmm. I can try that and see how it goes. \textcolor{blue}{[Preparation, Accept, neutral]} \\
\bottomrule
\end{tabular}
\caption{Complete dialogue for a generated conversation from MASC(Hetero), psychological state, communicative action, and emotion expression transition case.}
\label{tab:case-ps-ca-ee-transition-ours}
\end{table*}

\end{document}